\documentclass{article} 
\usepackage{iclr2019_conference,times}

\usepackage{amsmath,amsfonts,bm}

\def\eqref#1{equation~\ref{#1}}

\def\1{\bm{1}}

\def\rvh{{\mathbf{h}}}

\def\rvm{{\mathbf{m}}}

\def\rvr{{\mathbf{r}}}

\def\rvt{{\mathbf{t}}}

\def\rvv{{\mathbf{v}}}

\def\rvx{{\mathbf{x}}}
\def\rvy{{\mathbf{y}}}
\def\rvz{{\mathbf{z}}}

\def\mW{{\bm{W}}}

\DeclareMathAlphabet{\mathsfit}{\encodingdefault}{\sfdefault}{m}{sl}
\SetMathAlphabet{\mathsfit}{bold}{\encodingdefault}{\sfdefault}{bx}{n}

\DeclareMathOperator*{\argmax}{arg\,max}

\usepackage{hyperref}
\usepackage{url}
\usepackage{array}
\usepackage{amsthm}
\usepackage{graphicx}
\usepackage{subcaption}

\usepackage{pifont}
\newcommand{\cmark}{\ding{51}}
\newcommand{\xmark}{\ding{55}}

\title{RotatE: Knowledge Graph Embedding by Relational Rotation in Complex Space}

\author{
Zhiqing Sun $^1$\thanks{This work was done when the first author was visiting Mila and Universit\'e de Montr\'eal.}\ , Zhi-Hong Deng$^1$, Jian-Yun Nie$^3$, Jian Tang$^{2,4,5}$\\
$^1$Peking University, China\\
$^2$Mila-Quebec Institute for Learning Algorithms, Canada\\
$^3$Universit\'e de Montr\'eal, Canada\\
$^4$HEC Montr\'eal, Canada\\
$^5$CIFAR AI Research Chair\\
\texttt{\{1500012783, zhdeng\}@pku.edu.cn}\\
\texttt{nie@iro.umontreal.ca}\\
\texttt{jian.tang@hec.ca}
}

\usepackage{amsthm}
\newtheorem{theorem}{Theorem}

\usepackage{physics}

\renewcommand{\norm}[1]{\left\lVert#1\right\rVert}
\def\method{RotatE}
\def\baseline{pRotatE}
\def\tth{{\mathtt{h}}}
\def\ttr{{\mathtt{r}}}
\def\ttt{{\mathtt{t}}}
\def\ttx{{\mathtt{x}}}
\def\tty{{\mathtt{y}}}
\def\ttz{{\mathtt{z}}}

\DeclareSymbolFont{extraup}{U}{zavm}{m}{n}
\DeclareMathSymbol{\varheart}{\mathalpha}{extraup}{86}
\DeclareMathSymbol{\vardiamond}{\mathalpha}{extraup}{87}

\newtheorem{definition}{Definition}
\newtheorem{lemma}[theorem]{Lemma}

\iclrfinalcopy 
\begin{document}
\maketitle

\begin{abstract}
We study the problem of learning representations of entities and relations in knowledge graphs for predicting missing links. The success of such a task heavily relies on the ability of modeling and inferring the patterns of (or between) the relations. In this paper, we present a new approach for knowledge graph embedding called RotatE, which is able to model and infer various relation patterns including: symmetry/antisymmetry, inversion, and composition.
Specifically, the RotatE model defines each relation as a rotation from the source entity to the target entity in the complex vector space. In addition, we propose a novel self-adversarial negative sampling technique for efficiently and effectively training the RotatE model. Experimental results on multiple benchmark knowledge graphs show that the proposed RotatE model is not only scalable, but also able to infer and model various relation patterns and significantly outperform existing state-of-the-art models for link prediction.
\end{abstract}

\section{Introduction}

Knowledge graphs are collections of factual triplets, where each triplet $(\tth, \ttr, \ttt)$ represents a relation $\ttr$ between a head entity $\tth$ and a tail entity $\ttt$. Examples of real-world knowledge graphs include Freebase \citep{bollacker2008freebase}, Yago \citep{suchanek2007yago}, and WordNet \citep{miller1995wordnet}. Knowledge graphs are potentially useful to a variety of applications such as question-answering \citep{hao2017end}, information retrieval \citep{xiong2017explicit}, recommender systems \citep{zhang2016collaborative}, and natural language processing \citep{yang2017leveraging}. Research on knowledge graphs is attracting growing interests in both academia and industry communities. 

Since knowledge graphs are usually incomplete, a fundamental problem for knowledge graph is predicting the missing links. Recently, extensive studies have been done on learning low-dimensional representations of entities and relations for missing link prediction (a.k.a., knowledge graph embedding) \citep{bordes2013translating,trouillon2016complex,dettmers2017convolutional}.
These methods have been shown to be scalable and effective. The general intuition of these methods is to model and infer the connectivity patterns in knowledge graphs according to the observed knowledge facts. For example, some relations are symmetric (e.g., marriage) while others are antisymmetric (e.g., filiation); 
some relations are the inverse of other relations (e.g., hypernym and hyponym); 
and some relations may be composed by others (e.g., my mother's husband is my father). It is critical to find ways to model and infer these patterns, i.e., \textbf{symmetry/antisymmetry}, \textbf{inversion}, and \textbf{composition}, from the observed facts in order to predict missing links. 



Indeed, many existing approaches have been trying to either implicitly or explicitly model one or a few of the above relation patterns \citep{bordes2013translating,wang2014knowledge,lin2015learning,yang2014embedding,trouillon2016complex}. 
For example, the TransE model \citep{bordes2011learning}, which represents relations as translations, aims to model the inversion and composition patterns; the DisMult model \citep{yang2014embedding}, which models the three-way interactions between head entities, relations, and tail entities, aims to model the symmetry pattern. However, none of existing models is capable of modeling and inferring all the above patterns. Therefore, we are looking for an approach that is able to model and infer all the three types of relation patterns. 


In this paper, we propose such an approach called \method{} for knowledge graph embedding.
Our motivation is from Euler's identity $e^{i\theta} = \cos \theta + i \sin \theta$, which indicates that a unitary complex number can be regarded as a rotation in the complex plane.
Specifically, the \method{} model maps the entities and relations to the complex vector space and defines each relation as a rotation from the source entity to the target entity. Given a triplet $(\tth,\ttr,\ttt)$, we expect that $\rvt = \rvh \circ \rvr$, where $\rvh, \rvr, \rvt \in \mathbb{C}^k$ are the embeddings, the modulus $|r_i| =1$ and $\circ$ denotes the Hadamard (element-wise) product. Specifically, for each dimension in the complex space, we expect that:
\begin{equation}
    t_i = h_i r_i,  \text{ where }h_i, r_i, t_i \in \mathbb{C} \text{ and } |r_i| =1.
\end{equation}

It turns out that such a simple operation can effectively model all the three relation patterns: symmetric/antisymmetric, inversion, and composition. For example, a relation $\ttr$ is symmetric if and only if each element of its embedding $\rvr$, i.e. $r_i$, satisfies $r_i = e^{0 / i\pi} = \pm 1$; two relations $\ttr_1$ and $\ttr_2$ are inverse if and only if their embeddings are conjugates: $\rvr_2=\bar{\rvr}_1$; a relation $\rvr_3 = e^{i\boldsymbol{\theta_3}}$ is a combination of other two relations $\rvr_1 = e^{i\boldsymbol{\theta_1}}$ and $\rvr_2 = e^{i\boldsymbol{\theta_2}}$ if and only if $\rvr_3 = \rvr_1 \circ \rvr_2$ (i.e. $\boldsymbol{\theta_3} = \boldsymbol{\theta_1} + \boldsymbol{\theta_2}$). Moreover, the \method{} model is scalable to large knowledge graphs as it remains linear in both time and memory. 

To effectively optimizing the \method{}, we further propose a novel self-adversarial negative sampling technique, which generates negative samples according to the current entity and relation embeddings. The proposed technique is very general and can be applied to many existing knowledge graph embedding models. We evaluate the \method{} on four large knowledge graph benchmark datasets including FB15k \citep{bordes2013translating}, WN18 \citep{bordes2013translating}, FB15k-237 \citep{toutanova2015observed} and WN18RR \citep{dettmers2017convolutional}. Experimental results show that the \method{} model significantly outperforms existing state-of-the-art approaches. In addition, \method{} also outperforms state-of-the-art models on Countries \citep{bouchard2015approximate}, a benchmark explicitly designed for composition pattern inference and modeling. To the best of our knowledge, \method{} is the first model that achieves state-of-the-art performance on all the benchmarks.\footnote{The codes of our paper are available online: \url{https://github.com/DeepGraphLearning/KnowledgeGraphEmbedding}.}


\section{Related Work}

\begin{table}[t]
\centering
\small
\begin{tabular}{|c|c|c|}
\hline
\textbf{Model} &  \multicolumn{2}{c|}{\textbf{Score Function}}\\
\hline
\hline
SE \citep{bordes2011learning} & $-\norm{\mW_{r,1}\rvh - \mW_{r,2}\rvt}$ &
$\rvh, \rvt \in \mathbb{R}^k, \mW_{r,\cdot} \in  \mathbb{R}^{k \times k}$\\
\hline
TransE \citep{bordes2013translating} & $-\norm{\rvh + \rvr - \rvt}$ & $\rvh, \rvr, \rvt \in \mathbb{R}^k$\\
\hline
TransX & $-\norm{g_{r, 1}(\rvh) + \rvr - g_{r, 2}(\rvt)}$ & $\rvh, \rvr, \rvt \in \mathbb{R}^k$\\
\hline
DistMult \citep{yang2014embedding} & $ \langle \rvr, \rvh, \rvt \rangle$ & $\rvh, \rvr, \rvt \in \mathbb{R}^k$\\
\hline
ComplEx \citep{trouillon2016complex} & $ \Re(\langle \rvr, \rvh, \overline{\rvt} \rangle)$ & $\rvh, \rvr, \rvt \in \mathbb{C}^k$\\
\hline
HolE \citep{nickel2016holographic} & $ \langle \rvr, \rvh \otimes \rvt \rangle$ & $\rvh, \rvr, \rvt \in \mathbb{R}^k$\\
\hline
ConvE \citep{dettmers2017convolutional} & $ \langle \sigma(\mathrm{vec}(\sigma([ \overline{\rvr\vphantom{h}}, \overline{\rvh}] \ast \boldsymbol{\Omega})) \mW), \rvt \rangle$ & $\rvh, \rvr, \rvt \in \mathbb{R}^k$\\
\hline
\hline
\method{} & $-\norm{\rvh \circ \rvr - \rvt}$\footnotemark & $\rvh, \rvr, \rvt \in \mathbb{C}^k, |r_i| = 1$\\
\hline
\end{tabular}
\caption{The score functions $f_r(\rvh, \rvt)$ of several knowledge graph embedding models, where $\langle \cdot \rangle$ denotes the generalized dot product, $\circ$ denotes the Hadamard product, $\otimes$ denotes circular correlation, $\sigma$ denotes activation function and $\ast$ denotes 2D convolution. $\overline{\ \cdot\ }$ denotes conjugate for complex vectors, and 2D reshaping for real vectors in ConvE model. TransX represents a wide range of TransE's variants, such as TransH  \citep{wang2014knowledge}, TransR \citep{lin2015learning}, and STransE \citep{nguyen2016stranse}, where $g_{r, i}(\cdot)$ denotes a matrix multiplication with respect to relation $\ttr$.}
\label{tab:score}
\end{table}

\footnotetext{The $p$-norm of a complex vector $\rvv$ is defined as $\norm{\rvv}_p = \sqrt[p]{\sum |\rvv_i|^p}$. We use L1-norm for all distance-based models in this paper and drop the subscript of $\norm{\cdot}_1$ for brevity.}

Predicting missing links with \textbf{knowledge graph embedding} (KGE) methods has been extensively investigated in recent years. The general methodology is to define a score function for the triplets. Formally, let $\mathcal{E}$ denote the set of entities and $\mathcal{R}$ denote the set of relations, then a knowledge graph is a collection of factual triplets $(\tth, \ttr, \ttt)$, where $\tth, \ttt \in \mathcal{E}$ and $\ttr \in \mathcal{R}$. Since entity embeddings are usually represented as vectors, the score function usually takes the form $f_r(\rvh, \rvt)$, where $\rvh$ and $\rvt$ are head and tail entity embeddings. The score function $f_r(\rvh, \rvt)$ measures the salience of a candidate triplet $(\tth, \ttr, \ttt)$. The goal of the optimization is usually to score true triplet $(\tth, \ttr, \ttt)$ higher than the corrupted false triplets $(\tth', \ttr, \ttt)$ or $(\tth, \ttr, \ttt')$. Table \ref{tab:score} summarizes different score functions $f_r(\rvh, \rvt)$ in previous state-of-the-art methods as well as the model proposed in this paper. 
These models generally capture only a portion of the relation patterns. For example, TransE represents each relation as a bijection between source entities and target entities, and thus implicitly models inversion and composition of relations, but it cannot model symmetric relations; ComplEx extends DistMult by introducing complex embeddings so as to better model asymmetric relations, but it cannot infer the composition pattern.
The proposed \method{} model leverages the advantages of both.

A relevant and concurrent work to our work is the TorusE \citep{ebisu2018toruse} model, which defines knowledge graph embedding as translations on a compact Lie group. The TorusE model can be regarded as a special case of RotatE, where the modulus of embeddings are set fixed; our RotatE is defined on the entire complex space, which has much more representation capacity. Our experiments show that this is very critical for modeling and inferring the composition patterns. Moreover, TorusE focuses on the problem of regularization in TransE while this paper focuses on modeling and inferring multiple types of relation patterns.

There are also a large body of relational approaches for modeling the relational patterns on knowledge graphs \citep{lao2011random,neelakantan2015compositional,das2016chains,rocktaschel2017end,yang2017differentiable}. However, these approaches mainly focus on explicitly modeling the relational paths while our proposed RotatE model implicitly learns the relation patterns, which is not only much more scalable but also provides meaningful embeddings for both entities and relations. 


Another related problem is how to effectively draw negative samples for training knowledge graph embeddings. This problem has been explicitly studied by \cite{cai2017kbgan}, which proposed a generative adversarial learning framework to draw negative samples. However, such a framework requires simultaneously training the embedding model and a discrete negative sample generator, which are difficult to optimize and also computationally expensive. We propose a self-adversarial sampling scheme which only relies on the current model. It does require any additional optimization component, which make it much more efficient.

\section{\method{}: Relational Rotation in Complex Vector Space}
In this section, we introduce our proposed \method{} model. We first introduce three important relation patterns that are widely studied in the literature of link prediction on knowledge graphs. Afterwards, we introduce our proposed \method{} model, which defines relations as rotations in complex vector space. We also show that the \method{} model is able to model and infer all three relation patterns. 

\begin{table}[t]
\small
\centering
\begin{tabular}{|c|c|c|c|c|c|}
\hline
\textbf{Model} &  \textbf{Score Function} & \textbf{Symmetry}  & \textbf{Antisymmetry} & \textbf{Inversion} & \textbf{Composition}\\
\hline
\hline
SE & $-\norm{\mW_{r,1}\rvh - \mW_{r,2}\rvt}$ & \xmark & \xmark & \xmark & \xmark\\
\hline
TransE & $-\norm{\rvh + \rvr - \rvt}$& \xmark & \cmark & \cmark & \cmark\\
\hline
TransX & $-\norm{g_{r, 1}(\rvh) + \rvr - g_{r, 2}(\rvt)}$& \cmark & \cmark & \xmark & \xmark\\
\hline
DistMult & $ \langle \rvh, \rvr, \rvt \rangle$& \cmark & \xmark & \xmark & \xmark\\
\hline
ComplEx & $ \Re(\langle \rvh, \rvr, \overline{\rvt} \rangle)$& \cmark & \cmark & \cmark & \xmark\\
\hline
\hline
\method{} & $-\norm{\rvh \circ \rvr - \rvt}$ & \cmark & \cmark & \cmark & \cmark\\
\hline
\end{tabular}
\caption{The pattern modeling and inference abilities of several models.}
\label{tab:modeling}
\end{table}

\subsection{Modeling and Inferring Relation Patterns}


The key of link prediction in knowledge graph
is to infer the connection patterns, e.g., relation patterns, with observed facts. According to the existing literature \citep{trouillon2016complex,toutanova2015observed,guu2015traversing,lin2015modeling}, three types of relation patterns are very important and widely spread in knowledge graphs: symmetry, inversion and composition. We give their formal definition here:

\begin{definition}
A relation $\ttr$ is \textbf{symmetric} (\textbf{antisymmetric}) if $\forall \ttx, \tty$
\begin{equation*}
	\ttr(\ttx, \tty) \Rightarrow \ttr(\tty, \ttx)\ \left(\ \ttr(\ttx, \tty) \Rightarrow \neg \ttr(\tty,\ttx) \ \right)
\end{equation*}
A clause with such form is a \textbf{symmetry (antisymmetry)} pattern.
\end{definition}

\begin{definition}
Relation $\ttr_1$ is \textbf{inverse} to relation $\ttr_2$ if $\forall \ttx, \tty$
\begin{equation*}
	\ttr_2 (\ttx, \tty) \Rightarrow \ttr_1 (\tty, \ttx)
\end{equation*}
A clause with such form is a \textbf{inversion} pattern.
\end{definition}

\begin{definition}
Relation $\ttr_1$ is \textbf{composed} of relation $\ttr_2$ and relation $\ttr_3$ if $\forall \ttx, \tty, \ttz$
\begin{equation*}
	\ttr_2 (\ttx, \tty) \wedge \ttr_3 (\tty, \ttz) \Rightarrow \ttr_1 (\ttx, \ttz)
\end{equation*}
A clause with such form is a \textbf{composition} pattern.
\end{definition}

According to the definition of the above three types of relation patterns, we provide an analysis of existing models on their abilities in inferring and modeling these patterns. Specifically, we provide an analysis on TransE, TransX, DistMult, and ComplEx.\footnote{See discussion at Appendix \ref{sec:dis}} We did not include the analysis on HolE and ConvE since HolE is equivalent to ComplEx  \citep{hayashi2017equivalence}, and ConvE is a black box that involves two-layer neural networks and convolution operations, which are hard to analyze. The results are summarized into Table \ref{tab:modeling}.
We can see that no existing approaches are capable of modeling all the three relation patterns. 

\subsection{Modeling Relations as Rotations in Complex Vector Space}

\begin{figure}[t]
\centering
\begin{subfigure}{.3\textwidth}
  \centering
  \captionsetup{width=.8\linewidth}
  \includegraphics[width=0.8\linewidth]{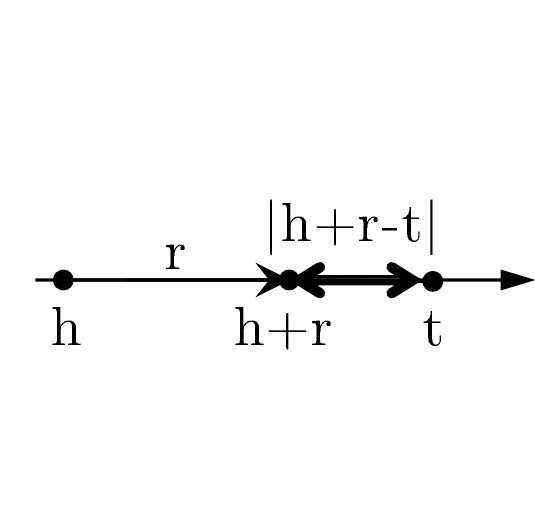}
  \caption{TransE models $\ttr$ as translation in real line.}
  \label{fig:1}
\end{subfigure}%
\begin{subfigure}{.3\textwidth}
  \centering
  \captionsetup{width=.8\linewidth}
  \includegraphics[width=0.8\linewidth]{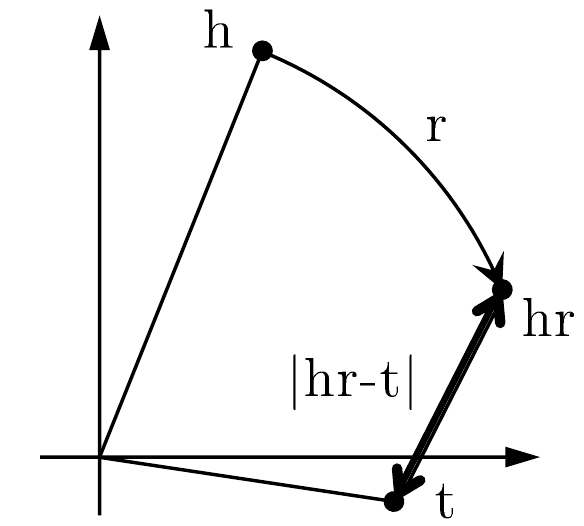}
  \caption{\method{} models $\ttr$ as rotation in complex plane.}
  \label{fig:4}
\end{subfigure}
\begin{subfigure}{.3\textwidth}
  \centering
  \captionsetup{width=.8\linewidth}
  \includegraphics[width=0.8\linewidth]{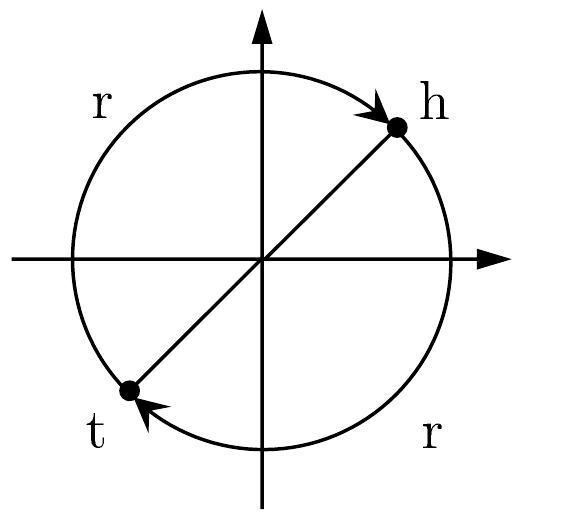}
  \caption{RotatE: an example of modeling symmetric relations $\ttr$ with $r_i=-1$}
  \label{fig:3}
\end{subfigure}
\caption{Illustrations of TransE and \method{} with only 1 dimension of embedding.}
\label{fig:fig}
\end{figure}

In this part, we introduce our proposed model that is able to model and infer all the three types of relation patterns. Inspired by Euler's identity, we map the head and tail entities $\tth, \ttt$ to the complex embeddings, i.e., $\rvh, \rvt \in \mathbb{C}^k$;
then we define the functional mapping induced by each relation $\ttr$ as  an element-wise rotation from the head entity $\rvh$ to the tail entity $\rvt$. In other words, given a triple $(\tth, \ttr, \ttt)$, we expect that:
%
\begin{equation}
\rvt = \rvh \circ \rvr\\, \text{\:\: where } |r_i| = 1,
\end{equation}

and $\circ$ is the Hadmard (or element-wise) product. Specifically, for each element in the embeddings, we have $t_i = h_i r_i$.
Here, we constrain the modulus of each element of $\rvr \in \mathbb{C}^k$, i.e., $r_i\in \mathbb{C}$, to be $|r_i|=1$. By doing this, $r_i$ is of the form $e^{i\theta_{r,i}}$, which corresponds to a counterclockwise rotation by $\theta_{r, i}$ radians about the origin of the complex plane, and only affects the phases of the entity embeddings in the complex vector space.
We refer to the proposed model as \method{} due to its rotational nature. According to the above definition, for each triple $(\tth, \ttr, \ttt)$, we define the distance function of \method{} as:
\begin{equation}
d_r(\rvh, \rvt) = \norm{\rvh \circ \rvr - \rvt}
\end{equation}

By defining each relation as a rotation in the complex vector spaces, \method{} can model and infer all the three types of relation patterns introduced above. Formally, we have following results\footnote{We relegate all proofs to the appendix.}:
\begin{lemma}\label{lem1}
\method{} can infer the symmetry/antisymmetry pattern. (See proof in Appendix \ref{sec:lem1})
\end{lemma}
\begin{lemma}\label{lem2}
\method{} can infer the inversion pattern. (See proof in Appendix \ref{sec:lem2})
\end{lemma}
\begin{lemma}\label{lem3}
\method{} can infer the composition pattern. (See proof in Appendix \ref{sec:lem3})
\end{lemma}
These results are also summarized into Table \ref{tab:modeling}.
We can see that the \method{} model is  the only model that can model and infer all the three types of relation patterns.

\paragraph{Connection to TransE.} From Table \ref{tab:modeling}, we can see that TransE is able to infer and model all the other relation patterns except the symmetry pattern. The reason is that in TransE, any symmetric relation will be represented by a $\mathbf{0}$ translation vector. As a result, this will push the entities with symmetric relations to be close to each other in the embedding space. \method{} solves this problem and is a able to model and infer the symmetry pattern. An arbitrary vector $\rvr$ that satisfies $r_i = \pm 1$ can be used for representing a symmetric relation in \method{}, and thus the entities having symmetric relations can be distinguished. Different symmetric relations can be also represented with different embedding vectors.  Figure \ref{fig:fig} provides illustrations of TransE and \method{} with only 1-dimensional embedding and shows how \method{} models a symmetric relation.
%
%

\subsection{Optimization}



Negative sampling has been proved quite effective for both learning knowledge graph embedding \citep{trouillon2016complex} and word embedding \citep{mikolov2013distributed}. Here we use a loss function similar to the negative sampling loss \citep{mikolov2013distributed} for effectively optimizing distance-based models:
\begin{equation}
L = - \log \sigma(\gamma - d_r(\rvh, \rvt)) - \sum_{i=1}^n \frac{1}{k} \log \sigma(d_r(\rvh'_i, \rvt'_i) - \gamma),
\end{equation}
where $\gamma$ is a fixed margin, $\sigma$ is the sigmoid function, and $(\tth_i', \ttr, \ttt_i')$ is the $i$-th negative triplet.

We also propose a new approach for drawing negative samples. The negative sampling loss samples the negative triplets in a uniform way. Such a uniform negative sampling suffers the problem of inefficiency since many samples are obviously false as training goes on, which does not provide any meaningful information. Therefore, we propose an approach called self-adversarial negative sampling, which samples negative triples according to the current embedding model. Specifically, we sample negative triples from the following distribution:
\begin{equation}
 p(h'_j, r, t'_j | \{(h_i, r_i, t_i)\}) = \frac{\exp \alpha f_r(\rvh'_j, \rvt'_j)}{\sum_i \exp \alpha f_r(\rvh'_i, \rvt'_i)}
\end{equation}
where $\alpha$ is the temperature of sampling. Moreover, since the sampling procedure may be costly, we treat the above probability as the weight of the negative sample. Therefore, the final negative sampling loss with self-adversarial training takes the following form:
\begin{equation}
L = - \log \sigma(\gamma - d_r(\rvh, \rvt)) - \sum_{i=1}^n p(h'_i, r, t'_i) \log \sigma(d_r(\rvh'_i, \rvt'_i) - \gamma)
\end{equation}


In the experiments, we will compare different approaches for negative sampling.

\section{Experiments}
\subsection{Experimental Setting}

\begin{table}[t]
\centering
\small
\begin{tabular}{|c|c|c|c|c|c|}
\hline
\textbf{Dataset} & \textbf{\#entity} & \textbf{\#relation} & \textbf{\#training} & \textbf{\#validation} & \textbf{\#test}\\
\hline
FB15k & 14,951 & 1,345 & 483,142 & 50,000 & 59,071\\
\hline
WN18 & 40,943 & 18 & 141,442 & 5,000 & 5,000\\
\hline
FB15k-237 & 14,541 & 237 & 272,115 & 17,535 & 20,466\\
\hline
WN18RR & 40,943 & 11 & 86,835 & 3,034 & 3,134\\
\hline
\end{tabular}
\caption{Number of entities, relations, and observed triples in each split for four benchmarks.}
\label{tab:meta}
\end{table}

We evaluate our proposed model on four widely used knowledge graphs. The statistics of these knowledge graphs are summarized into Table \ref{tab:meta}.
\begin{itemize}
\item FB15k  \citep{bordes2013translating} is a subset of Freebase  \citep{bollacker2008freebase}, a large-scale knowledge graph containing general knowledge facts. \cite{toutanova2015observed} showed that almost $81\%$  of the test triplets $\mathtt{(x, r, y)}$ can be inferred via a directly linked triplet $\mathtt{(x, r', y)}$ or $\mathtt{(y, r', x)}$. Therefore, the key of link prediction on FB15k is to model and infer the \textbf{symmetry/antisymmetry} and \textbf{inversion} patterns.

\item WN18  \citep{bordes2013translating} is a subset of WordNet  \citep{miller1995wordnet}, a database featuring lexical relations between words. This dataset also has many inverse relations. So the main relation patterns in WN18 are also \textbf{symmetry/antisymmetry} and \textbf{inversion}.

\item FB15k-237  \citep{toutanova2015observed} is a subset of FB15k, where inverse relations are deleted. Therefore, the key of link prediction on FB15k-237 boils down to model and infer the \textbf{symmetry/antisymmetry} and \textbf{composition} patterns.

\item WN18RR  \citep{dettmers2017convolutional} is a subset of WN18. The inverse relations are deleted, and the main relation patterns are \textbf{symmetry/antisymmetry} and \textbf{composition}.
\end{itemize}


\paragraph{Hyperparameter Settings.} We use Adam  \citep{kingma2014adam} as the optimizer and fine-tune the hyperparameters on the validation dataset. The ranges of the hyperparameters for the grid search are set as follows: embedding dimension\footnote{Following \cite{trouillon2016complex}, we treat complex number as the same as real number with regard to the embedding dimension. If the same number of dimension is used for both the real and imaginary parts of the complex number as the real number, the number of parameters for the complex embedding would be twice the number of parameters for the embeddings in the real space.} $k \in \{125, 250, 500, 1000\}$, batch size $b \in \{512, 1024, 2048\}$, self-adversarial sampling temperature $\alpha \in \{0.5, 1.0\}$, and fixed margin $\gamma \in \{3, 6, 9, 12, 18, 24, 30\}$. Both the real and imaginary parts of the entity embeddings are uniformly initialized, and the phases of the relation embeddings are uniformly initialized between $0$ and $2\pi$. No regularization is used since we find that the fixed margin $\gamma$ could prevent our model from over-fitting. 

\paragraph{Evaluation Settings.} We evaluate the performance of link prediction in the filtered setting: we rank test triples against all other candidate triples not appearing in the training, validation, or test set, where candidates are generated by corrupting subjects or objects: $(\tth', \ttr, \ttt)$ or $(\tth, \ttr, \ttt')$. Mean Rank (MR), Mean Reciprocal Rank (MRR) and Hits at N (H@N) are standard evaluation measures for these datasets and are evaluated in our experiments.

\paragraph{Baseline.} Apart from \method{}, we propose a variant of \method{} as baseline, where the modulus of the entity embeddings are also constrained: $|h_i| = |t_i| = C$, and the distance function is thus $2C\norm{\sin\frac{\boldsymbol{\theta}_h+\boldsymbol{\theta}_r-\boldsymbol{\theta}_t}{2}}$ (See Equation \ref{eqa:baseline} at Appendix \ref{sec:the1} for a detailed derivation). In this way, we can investigate how \method{} works without modulus information and with only phase information. We refer to the baseline as \baseline{}. It is obvious to see that \baseline{} can also model and infer all the three relation patterns.

\begin{table}[t]
\centering
\small
\begin{tabular}{|c|c c c c c|c c c c c|}
\hline
& \multicolumn{5}{c|}{\textbf{FB15k}} & \multicolumn{5}{c|}{\textbf{WN18}}\\
\cline{2-11}
& MR & MRR & H@1 & H@3 & H@10 & MR & MRR & H@1 & H@3 & H@10 \\
\hline
TransE [$\varheart$] & - & .463 & .297 & .578 & .749 & - & .495 & .113 & .888 & .943\\
\hline
DistMult [$\vardiamond$] & 42 & .798 & - & - & \textbf{.893} & 655 & .797 & - & - & .946\\
\hline
HolE & - & .524 & .402 & .613 & .739 & - & .938 & .930 & .945 & .949\\
\hline
ComplEx & - & .692 & .599 & .759 & .840 & - & .941 & .936 & .945 & .947\\
\hline
ConvE & 51 & .657 & .558 & .723 & .831 & 374 &  .943 & .935 & .946 & .956\\
\hline
\hline
\baseline{} & 43 & \textbf{.799} & \textbf{.750} & .829 & .884 & \textbf{254} & .947 & .942 & .950 & .957\\
\hline
\method{} & \textbf{40} & .797 & .746 & \textbf{.830} & .884 & 309 & \textbf{.949} & \textbf{.944} & \textbf{.952} & \textbf{.959}\\
\hline
\end{tabular}
\caption{Results of several models evaluated on the FB15K and WN18 datasets. Results of [$\varheart$] are taken from  \citep{nickel2016holographic} and results of [$\vardiamond$] are taken from  \citep{kadlec2017knowledge}. Other results are taken from the corresponding original papers.}
\label{tab:main1}
\end{table}

\begin{table}[t]
\small
\centering
\begin{tabular}{|c|c c c c c|c c c c c|}
\hline
& \multicolumn{5}{c|}{\textbf{FB15k-237}} & \multicolumn{5}{c|}{\textbf{WN18RR}}\\
\cline{2-11}
& MR & MRR & H@1 & H@3 & H@10 & MR & MRR & H@1 & H@3 & H@10 \\
\hline
TransE [$\varheart$] & 357 & .294 & - & - & .465 & 3384 & .226 & - & - & .501\\
\hline
DistMult & 254 & .241 & .155 & .263 & .419 & 5110 & .43 & .39 & .44 & .49\\
\hline
ComplEx & 339 & .247 & .158 & .275 & .428 & 5261 & .44 & .41 & .46 & .51\\
\hline
ConvE & 244 & .325 & .237 & .356 & .501 & 4187 & .43 & .40 & .44 & .52 \\
\hline
\hline
\baseline{} & 178 & .328 & .230 & .365 & .524 & \textbf{2923} & .462 & .417 & .479 & .552\\
\hline
\method{} & \textbf{177} & \textbf{.338} & \textbf{.241} & \textbf{.375} & \textbf{.533} & 3340 & \textbf{.476} & \textbf{.428} & \textbf{.492} & \textbf{.571}\\
\hline
\end{tabular}
\caption{Results of several models evaluated on the FB15k-237 and WN18RR datasets. Results of [$\varheart$] are taken from  \citep{nguyen2017novel}. Other results are taken from  \citep{dettmers2017convolutional}.}
\label{tab:main2}
\end{table}

\subsection{Main Results}

We compare \method{} to several state-of-the-art models, including TransE  \citep{bordes2013translating}, DistMult  \citep{yang2014embedding}, ComplEx  \citep{trouillon2016complex}, HolE  \citep{nickel2016holographic},
and ConvE  \citep{dettmers2017convolutional}, as well as our baseline model \baseline{}, to empirically show  the importance of modeling and inferring the relation patterns for the task of predicting missing links.

Table \ref{tab:main1} summarizes our results on FB15k and WN18. We can see that RotatE outperforms all the state-of-the-art models. The performance of \baseline{} and \method{} are similar on these two datasets. Table \ref{tab:main2} summarizes our results on FB15k-237 and WN18RR, where 
the improvement is much more significant. The difference between \method{} and \baseline{} is much larger on FB15k-237 and WN18RR, where there are a lot of composition patterns. This indicates that modulus is very important for modeling and inferring the composition pattern.

Moreover, the performance of these models on different datasets is consistent with our analysis on the three relation patterns (Table \ref{tab:modeling}):

\begin{itemize}
\item On FB15K, the main relation patterns are symmetry/antisymmetry and inversion. We can see that ComplEx performs well while TransE does not perform well since ComplEx can infer both symmetry/antisymmetry and inversion patterns while TransE cannot infer symmetry pattern. Surprisingly, DistMult achieves good performance on this dataset although it cannot model the antisymmetry and inversion patterns. The reason is that for most of the relations in FB15K, the types of head entities and tail entities are different. Although DistMult gives the same score to a true triplet $\mathtt{(h,r,t)}$ and its opposition triplet $\mathtt{(t,r,h)}$, $\mathtt{(t,r,h)}$ is usually impossible to be valid since the entity type of $t$ does not match the head entity type of $\tth$. For example, DistMult assigns the same score to $\mathtt{(Obama,nationality, USA)}$ and $\mathtt{(USA, nationality, Obama)}$. But $\mathtt{(USA, nationality, Obama)}$ can be simply predicted as false since $\mathtt{USA}$ cannot be the head entity of the relation $\mathtt{nationality}$. 



\item On WN18, the main relation patterns are also symmetry/antisymmetry and inversion. As expected, ComplEx still performs very well on this dataset. However, different from the results on FB15K, the performance of DistMult significantly decreases on WN18. The reason is that DistMult cannot model antisymmetry and inversion patterns, and almost all the entities in WN18 are words and belong to the same entity type, which do not have the same problem as FB15K. 



\item On FB15k-237, the main relation pattern is composition. We can see that TransE performs really well while ComplEx does not perform well. The reason is that, as discussed before, TransE is able to infer the composition pattern while ComplEx cannot infer the composition pattern.

\item On WN18RR, one of the main relation patterns is the symmetry pattern since almost each word has a symmetric relation in WN18RR, e.g., $also\_see$ and $similar\_to$. TransE does not well on this dataset since it is not able to model the symmetric relations. 
\end{itemize}
\vspace{-5mm}

\begin{table}[t]
\centering
\small
\begin{tabular}{|c|c|c|c|c|}
\hline
& \multicolumn{4}{c|}{\textbf{Countries} (AUC-PR)}\\
\cline{2-5}
& DistMult & ComplEx & ConvE & \method{} \\
\hline
S1 & $\mathbf{1.00\pm0.00}$ & $0.97\pm0.02$ & $\mathbf{1.00\pm0.00}$& $\mathbf{1.00\pm0.00}$\\
S2 & $0.72\pm0.12$ & $0.57\pm0.10$ & $0.99\pm0.01$& $\mathbf{1.00\pm0.00}$\\
S3 & $0.52\pm0.07$ & $0.43\pm0.07$ & $0.86\pm0.05$& $\mathbf{0.95\pm0.00}$\\
\hline
\end{tabular}
\caption{Results on the Countries datasets. Other results are taken from  \citep{dettmers2017convolutional}.}
\label{tab:auc}
\end{table}

\subsection{Inferring Relation Patterns on Countries DataSet}
We also evaluate our model on the Countries dataset \citep{bouchard2015approximate,nickel2016holographic}, which is carefully designed to explicitly test the capabilities of the link prediction models for composition pattern modeling and inferring. It contains 2 relations and 272 entities (244 countries, 5 regions and 23 subregions). Unlike link prediction on general knowledge graphs, the queries in Countries are of the form $\mathtt{locatedIn(c, ?)}$, and the answer is one of the five regions. The Countries dataset has 3 tasks, each requiring inferring a composition pattern with increasing length and difficulty.
For example, task S2 requires inferring a relatively simpler composition pattern:
\begin{equation*}
\mathtt{neighborOf(c_1, c_2) \wedge locatedIn(c_2, r) \Rightarrow locatedIn(c_1, r)},
\end{equation*}

while task S3 requires inferring the most complex composition pattern:
\begin{equation*}
\mathtt{neighborOf(c_1, c_2) \wedge locatedIn(c_2, s) \wedge locatedIn(s, r) \Rightarrow locatedIn(c_1, r)}.
\end{equation*}

In Table \ref{tab:auc}, we report the results with respect to the AUC-PR metric, which is commonly used in the literature. We can see that \method{} outperforms all the previous models. The performance of \method{}  is significantly better than other methods on S3, which is the most difficult task.

\begin{figure}[t]
\centering
\begin{subfigure}{0.32\textwidth}
  \centering
  \includegraphics[width=0.8\linewidth]{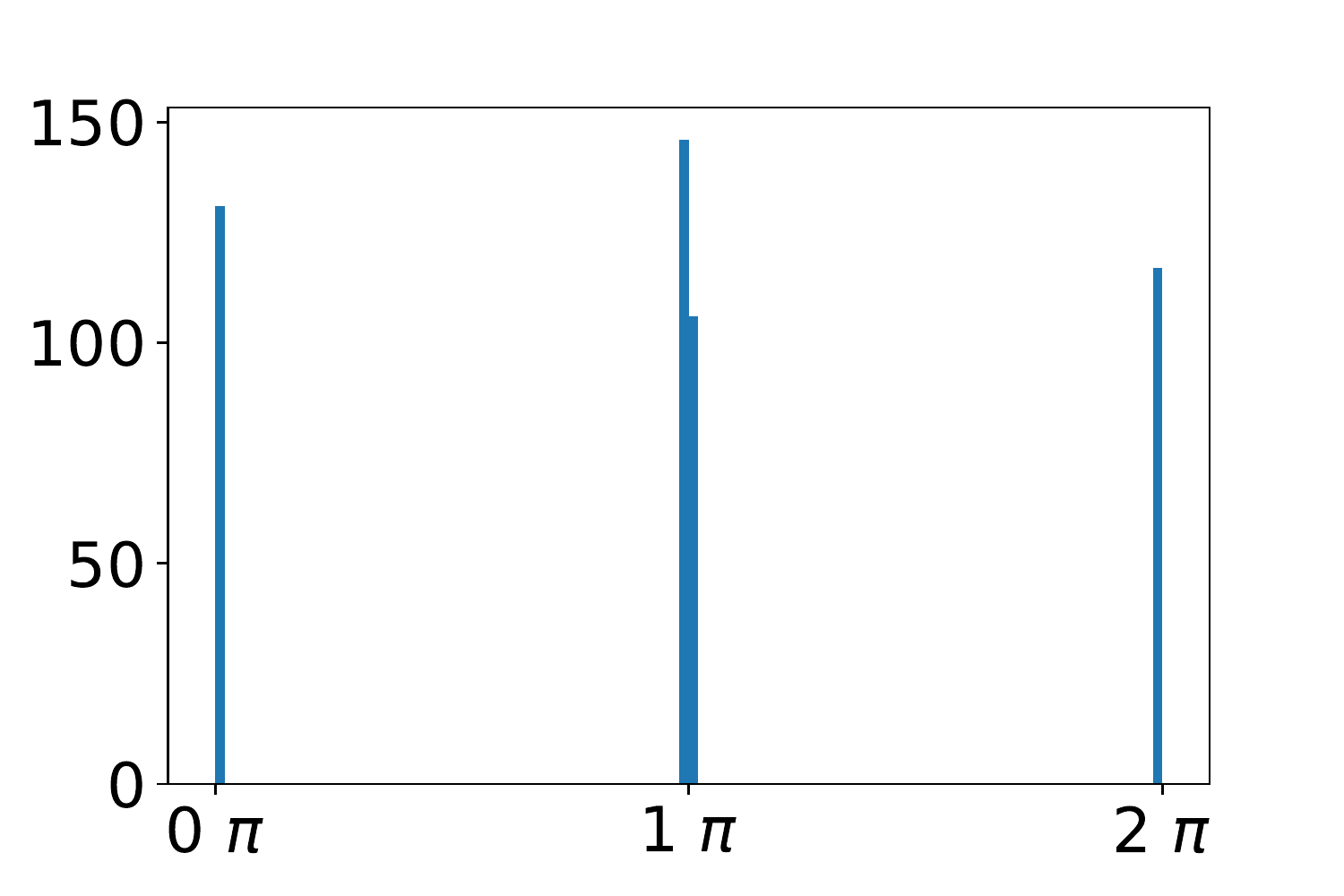}
  \caption{$\mathtt{similar\_to}$}
  \label{fig:sim}
\end{subfigure}
\begin{subfigure}{0.32\textwidth}
  \centering
  \includegraphics[width=0.8\linewidth]{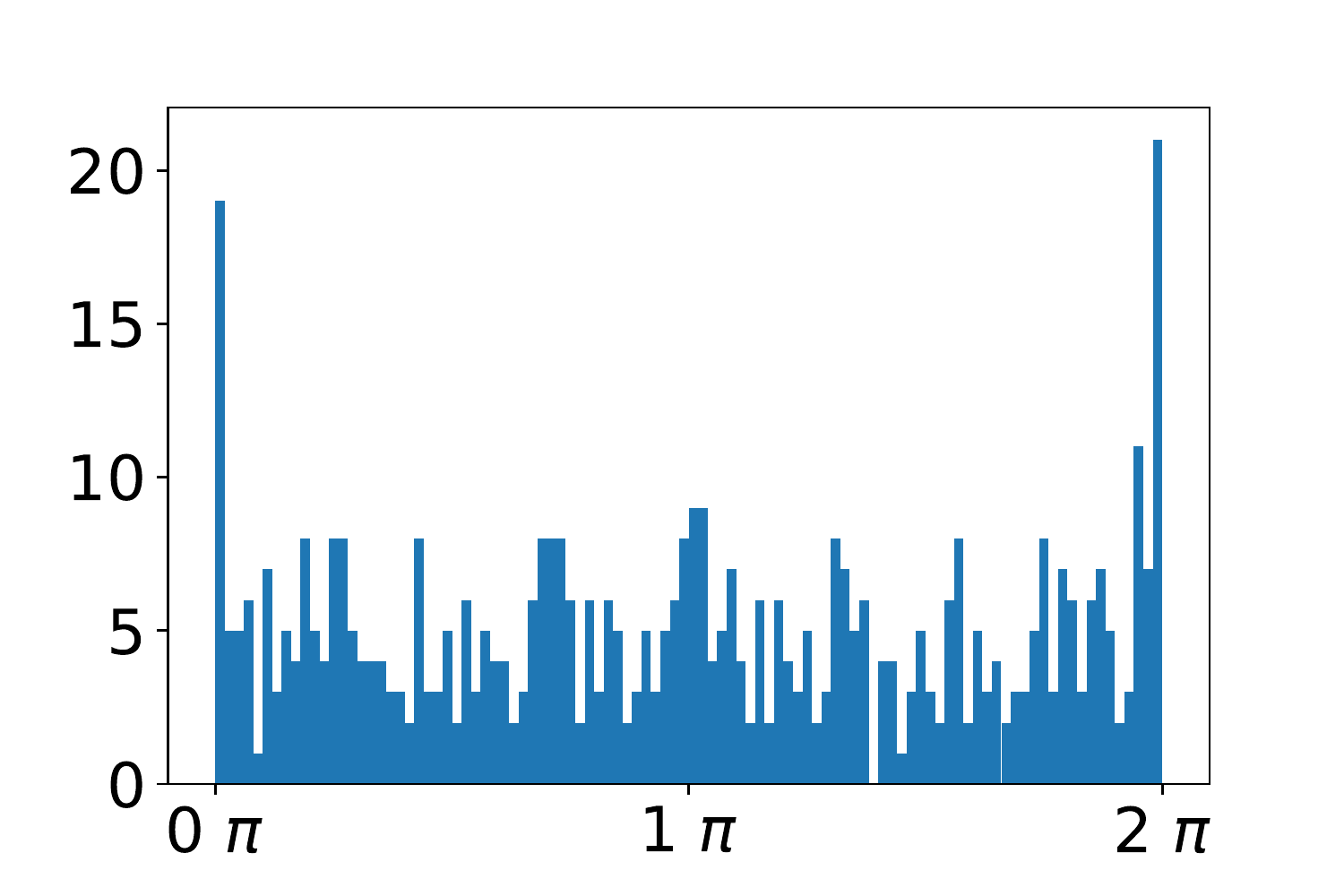}
  \caption{$\mathtt{hypernym}$}
  \label{fig:hyper}
\end{subfigure}
\begin{subfigure}{0.32\textwidth}
  \centering
  \includegraphics[width=0.8\linewidth]{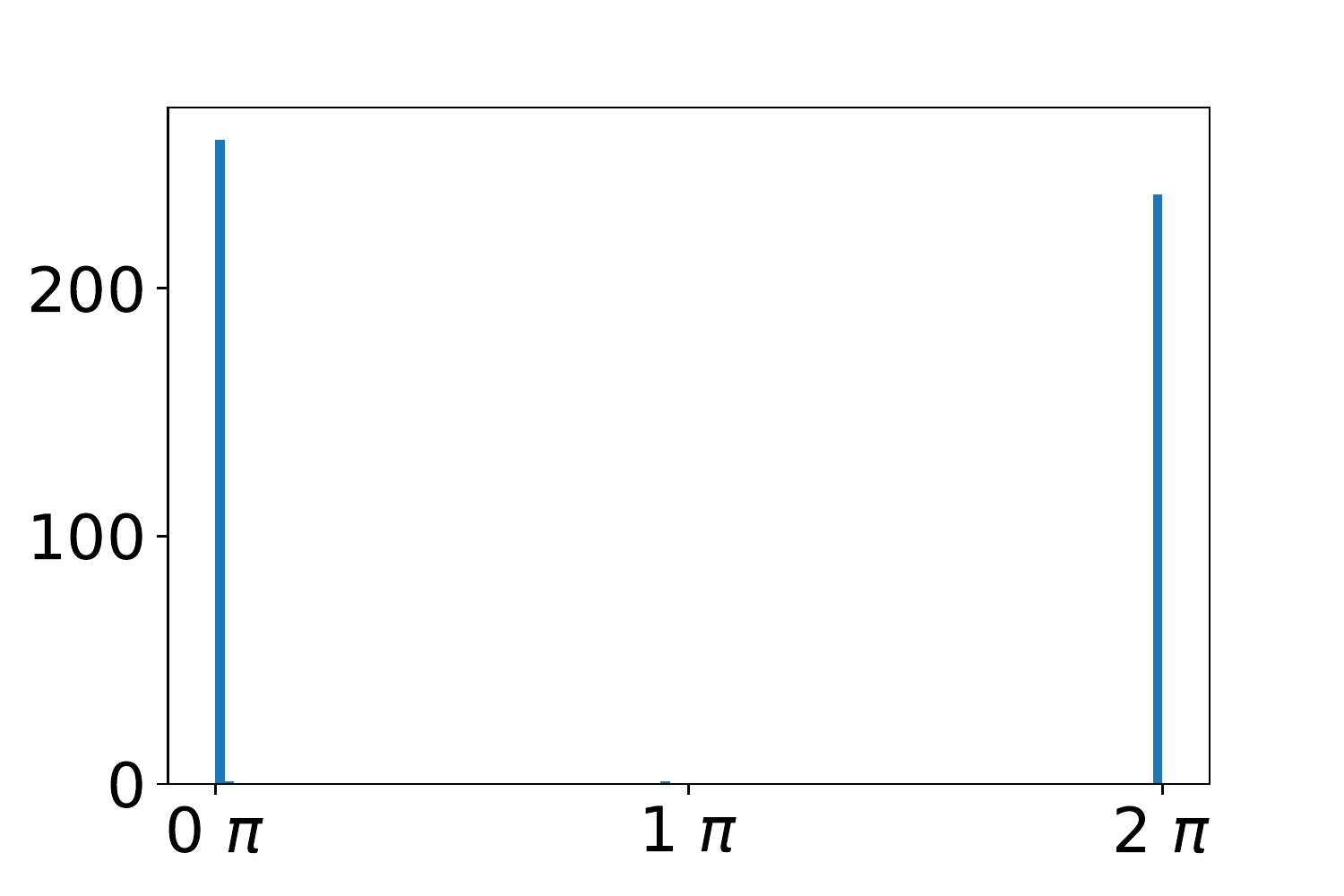}
  \caption{$\mathtt{hypernym \circ hyponym}$}
  \label{fig:hyper+hypo}
\end{subfigure}
\begin{subfigure}{0.24\textwidth}
  \centering
  \includegraphics[width=\linewidth]{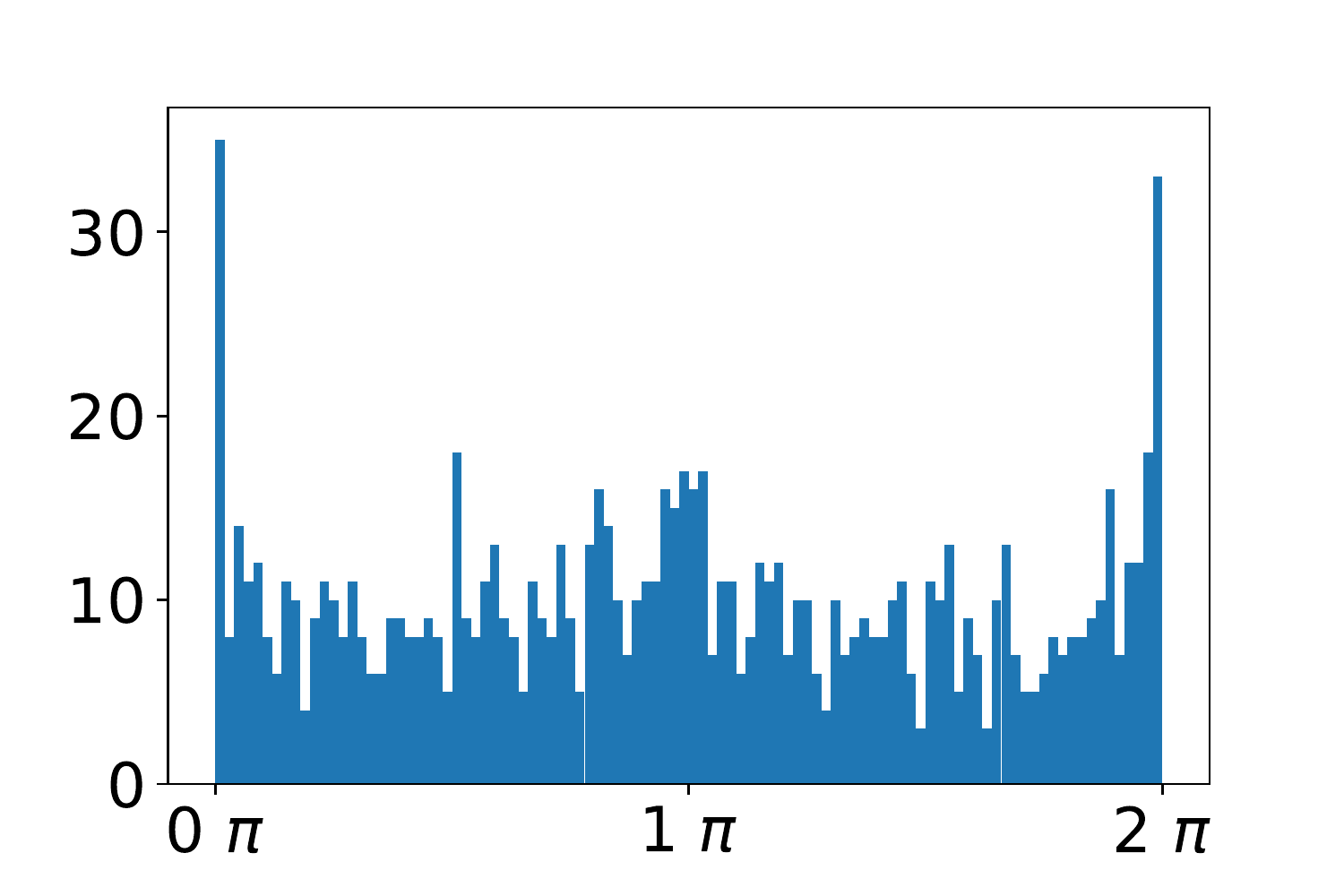}
  \caption{$\mathtt{for_1}$}
  \label{fig:nom}
\end{subfigure}
\begin{subfigure}{0.24\textwidth}
  \centering
  \includegraphics[width=\linewidth]{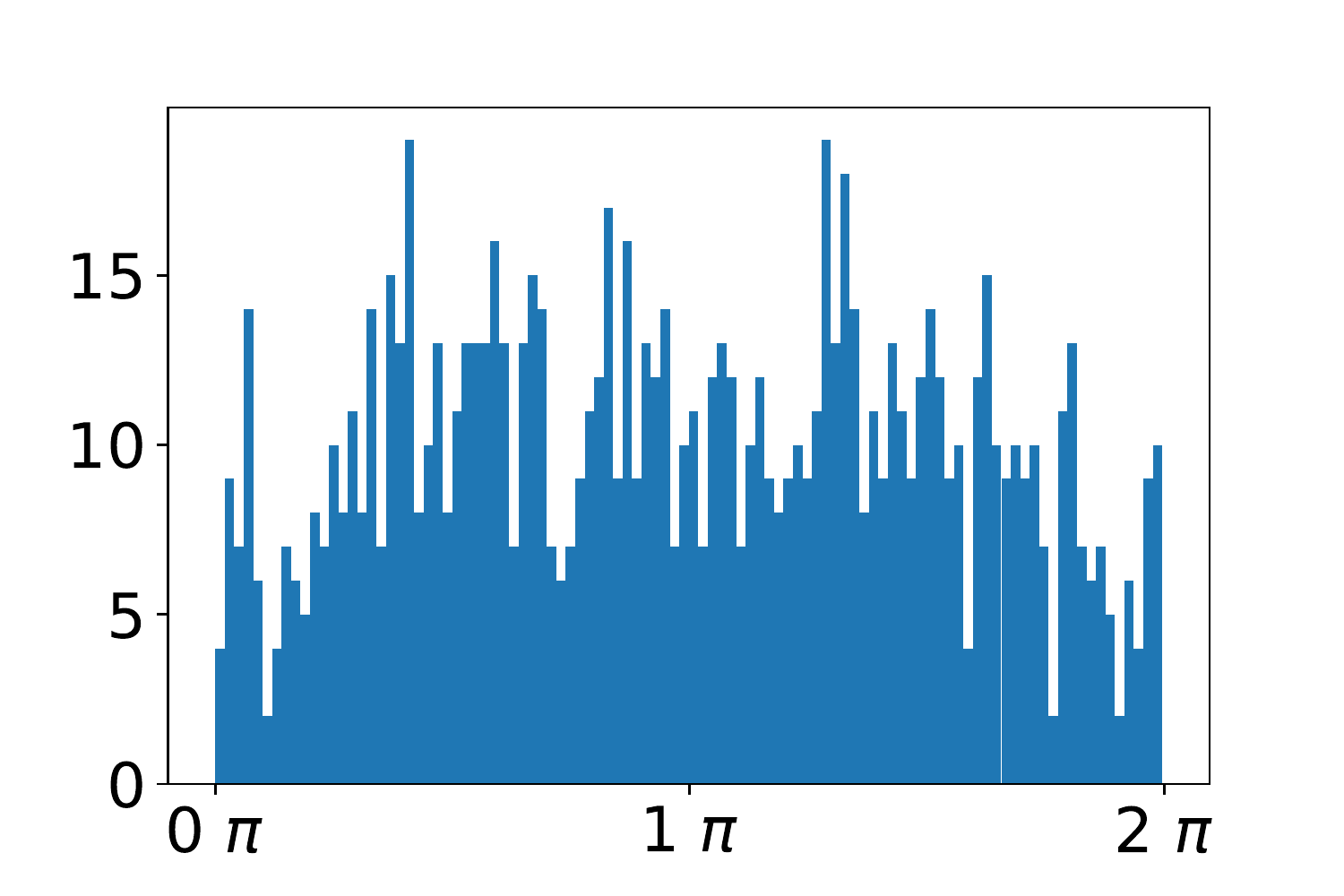}
  \caption{$\mathtt{winner}$}
  \label{fig:win}
\end{subfigure}
\begin{subfigure}{0.24\textwidth}
  \centering
  \includegraphics[width=\linewidth]{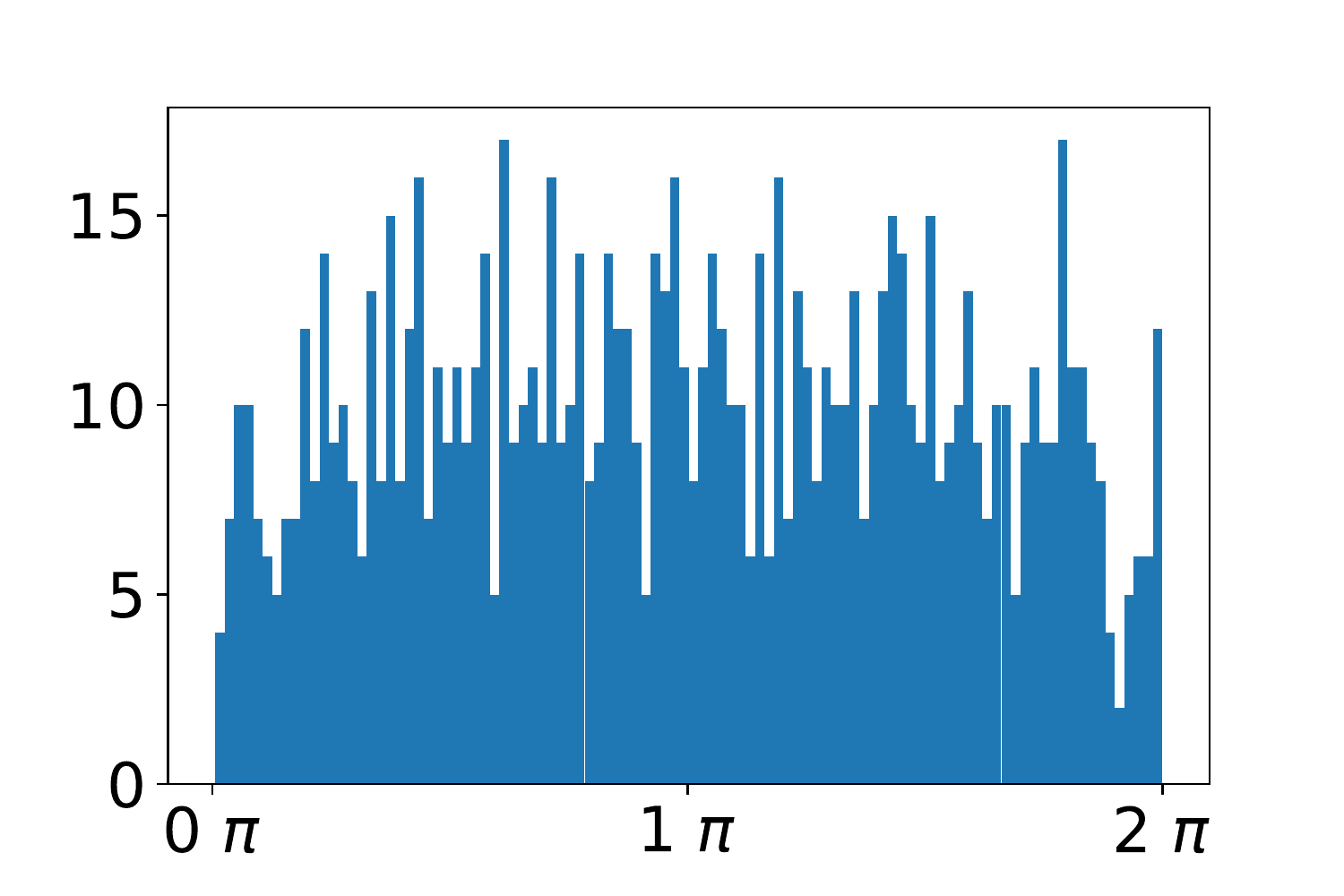}
  \caption{$\mathtt{for_2}$}
  \label{fig:nom2}
\end{subfigure}
\begin{subfigure}{0.24\textwidth}
  \centering
  \includegraphics[width=\linewidth]{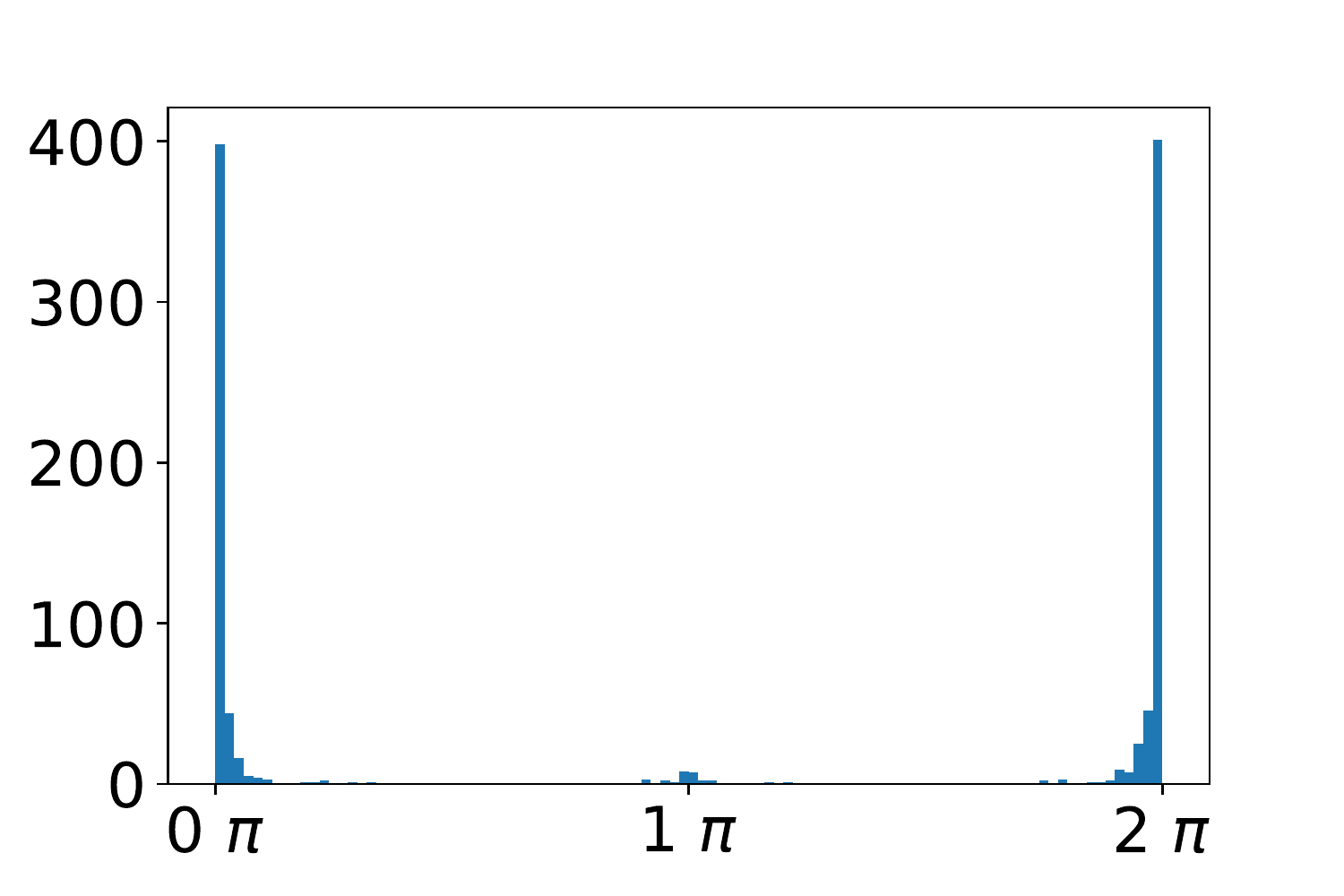}
  \caption{$\mathtt{for}_2^{-1} \circ \mathtt{winner}\circ \mathtt{for}_1$}
  \label{fig:nomw}
\end{subfigure}
\vspace{-2.0px}
\caption{Histograms of relation embedding phases $\{\theta_{r, i}\}$ $(r_i = e^{i\theta_{r, i}})$, where $\mathtt{for}_1$ represents relation $\mathtt{award\_nominee/award\_nominations./award/award\_nomination/nominated\_for}$, 
$\mathtt{winner}$ represents relation $\mathtt{award\_category/winners./award/award\_honor/award\_winner}$ and $\mathtt{for}_2$ represents $\mathtt{award\_category/nominees./award/award\_nomination/nominated\_for}$. The symmetry, inversion and composition pattern is represented in Figure \ref{fig:sim}, \ref{fig:hyper+hypo} and \ref{fig:nomw}, respectively.}
\label{fig:verify}
\end{figure}

\subsection{Implicit Relation Pattern Inference}

In this section, we verify whether the relation patterns are implicitly represented by \method{} relation embeddings. We ignore the specific positions in the relation embedding $\boldsymbol{\theta}_r$ and plot the histogram of the phase of each element in the relation embedding, i.e., \{$\theta_{r, i}$\}.

\textbf{Symmetry} pattern requires the symmetric relations to have property $\rvr\circ\rvr = \mathbf{1}$, and the solution is $r_i = \pm 1$. We investigate the relation embeddings from a $500$-dimensional \method{} trained on WN18. Figure \ref{fig:sim} gives the histogram of the embedding phases of a symmetric relation $similar\_to$. We can find that the embedding phases are either $\pi$ ($r_i = -1$) or $0,2\pi$ ($r_i = 1$). It indicates that the \method{} model does infer and model the symmetry pattern. Figure \ref{fig:hyper} is the histogram of relation $\mathtt{hypernym}$, which shows that the embedding of a general relation does not have such a $\pm 1$ pattern.

\textbf{Inversion} pattern requires the embeddings of a pair of inverse relations to be conjugate. We use the same \method{} model trained on WN18 for an analysis. Figure \ref{fig:hyper+hypo} illustrates the element-wise addition of the embedding phases from relation $\ttr_1 = \mathtt{hypernym}$ and its inversed relation $\ttr_2 = \mathtt{hyponym}$. All the additive embedding phases are $0$ or $2\pi$, which represents that $\rvr_1 = \rvr_2^{-1}$. This case shows that the inversion pattern is also inferred and modeled in the \method{} model.

\textbf{Composition} pattern requires the embedding phases of the composed relation to be the addition of the other two relations. Since there is no significant composition pattern in WN18, we study the inference of the composition patterns on FB15k-237, where a $1000$-dimensional \method{} is trained. Figure \ref{fig:nom} - \ref{fig:nomw} illustrate such a $\rvr_1 = \rvr_2 \circ \rvr_3$ case, where $\theta_{2,i} + \theta_{3,i} = \theta_{1,i}$ or  $\theta_{2,i} + \theta_{3,i} = \theta_{1,i} + 2\pi$. 

More results of implicitly inferring basic patterns are presented in the appendix.

\subsection{Comparing different negative sampling techniques}
In this part, we compare different negative sampling  techniques including uniform sampling, our proposed self-adversarial technique, and the KBGAN model \citep{cai2017kbgan}, which aims to
optimize a generative adversarial network to generate the negative samples. We re-implement a $50$-dimension TransE model with the margin-based ranking criterion that was used in \citep{cai2017kbgan}, and evaluate its performance on FB15k-237, WN18RR and WN18 with self-adversarial negative sampling. Table \ref{tab:neg} summarizes our results. We can see that self-adversarial sampling is the most effective negative sampling technique.

\begin{table}[t]
\centering
\begin{tabular}{|c|c c|c c|c c|}
\hline
& \multicolumn{2}{c|}{\textbf{FB15k-237}} & \multicolumn{2}{c|}{\textbf{WN18RR}}  & \multicolumn{2}{c|}{\textbf{WN18}}\\
\cline{2-7}
& MRR & H@10 & MRR & H@10 & MRR & H@10 \\
\hline
uniform & .242 & .422 & .186 & .459 & .433 & .915\\
\hline
KBGAN \citep{cai2017kbgan} & .278 & .453 & .210 & .479 & .705 & \textbf{.949}\\
\hline
\hline
self-adversarial & \textbf{.298} & \textbf{.475} & \textbf{.223} & \textbf{.510} & \textbf{.736} & .947\\
\hline
\end{tabular}
\caption{TransE with different negative sampling techniques. The results in first 2 rows are taken from \citep{cai2017kbgan}, where KBGAN uses a ComplEx negative sample generator.}
\label{tab:neg}
\end{table}

\begin{table}[t]
\centering
\begin{tabular}{|c|c c|c c|c c c|}
\hline
& \multicolumn{2}{c|}{\textbf{FB15k}} & \multicolumn{2}{c|}{\textbf{FB15k-237}} & \multicolumn{3}{c|}{\textbf{Countries (AUC-ROC)}}\\
\cline{2-8}
&  MRR & H@10 & MRR & H@10 & S1 & S2 & S3\\
\hline
TransE  & .735 & .871 & .332 & .531 & $\mathbf{1.00\pm0.00}$  & $\mathbf{1.00\pm0.00}$ & $\mathbf{0.96\pm0.00}$\\
\hline
ComplEx  & .780 & \textbf{.890}  & .319 & .509 & $\mathbf{1.00\pm0.00}$  & $0.98\pm0.00$ & $0.88\pm0.01$\\
\hline
\method{} & \textbf{.797} & .884  & \textbf{.338} & \textbf{.533} & $\mathbf{1.00\pm0.00}$  & $\mathbf{1.00\pm0.00}$ & $0.95\pm0.00$\\
\hline
\end{tabular}
\caption{Results of TransE and ComplEx with self-adversarial sampling and negative sampling loss on FB15k, FB15k-237 and Countries datasets.}
\label{tab:more}
\end{table}

\subsection{Further Experiments on TransE and ComplEx}
One may argue that the contribution of RotatE comes from the self-adversarial negative sampling technique. In this part, we conduct further experiments on TransE and ComplEx in the same setting as RotatE to make a fair comparison among the three models. Table \ref{tab:more} shows the results of TransE and ComplEx trained with the self-adversarial negative sampling technique on FB15k and FB15k-237 datasets, where a large number of relations are available. In addition, we evaluate these three models on the Countries dataset, which explicitly requires inferring the composition pattern. We also provide a detailed ablation study on TransE and RotatE in the appendix.

From Table \ref{tab:more}, we can see that similar results are observed as Table 4 and 5. The RotatE model achieves the best performance on both FB15k and FB15k-237, as it is able to model all the three relation patterns. The TransE model does not work well on the FB15k datasets, which requires modeling the symmetric pattern; the ComplEx model does not work well on FB15k-237, which requires modeling the composition pattern. The results on the Countries dataset are a little bit different, where the TransE model slightly outperforms RoateE on the S3 task. The reason is that the Countries datasets do not have the symmetric relation between different regions, and all the three tasks in the Countries datasets only require inferring the region for a given city. Therefore, the TransE model does not suffer from its inability of modeling symmetric relations. For ComplEx, we can see that it does not perform well on Countries since it cannot infer the composition pattern.

\begin{table}[t]
\centering
\small
\begin{tabular}{|c|c|c|c|c|c|c|c|c|}
\hline
Relation Category & 1-to-1 & 1-to-N & N-to-1 & N-to-N & 1-to-1 & 1-to-N & N-to-1 & N-to-N\\
\hline
\textbf{Tasks} & \multicolumn{4}{c|}{\textbf{Prediction Head (Hits@10)}} & \multicolumn{4}{c|}{\textbf{Prediction Tail (Hits@10)}}\\
\hline
TransE  & .437 & .657 & .182 & .472 & .437 & .197 & .667 & .500\\
\hline
TransH (bern)  & .668 & .876 & .287 & .645 & .655 & .398 & .833 & .672\\
\hline
KG2E\_KL (bern) & .923 & .946 & .660 & .696 & .926 & .679 & .944 & .734\\
\hline
\hline
TransE & .894 & \textbf{.972} & .567 & .880 & .879 & .671 & \textbf{.964} & .910\\
\hline
ComplEx & \textbf{.939} & .969 & \textbf{.692} & \textbf{.893} & \textbf{.938} & \textbf{.823} & .952 & .910\\
\hline
RotatE & .922 & .967 & .602 & \textbf{.893} & .923 & .713 & .961 & \textbf{.922}\\
\hline
\hline
\textbf{Tasks} & \multicolumn{4}{c|}{\textbf{Prediction Head (MRR)}} & \multicolumn{4}{c|}{\textbf{Prediction Tail (MRR)}}\\
\hline
TransE & .701 & .912 & .424 & .737 & .701 & .561 & .894 & .761\\
\hline
ComplEx & .832 & .914 & \textbf{.543} & .787 & .826 & \textbf{.661} & .869 & .800\\
\hline
RotatE & \textbf{.878} & \textbf{.934} & .465 & \textbf{.803} & \textbf{.872} & .611 & \textbf{.909} & \textbf{.832}\\
\hline
\end{tabular}
\caption{Experimental results on FB15k by relation category. The first three rows are taken from \citep{he2015learning}. The rest of the results are from RotatE trained with the self-adversarial negative sampling technique.}
\label{tab:one2one}
\end{table}

\subsection{Experimental results on FB15k by relation category}


We also did some further investigation on the performance of RotatE on different relation categories: one-to-many, many-to-one, and many-to-many relations\footnote{Following \cite{wang2014knowledge}, for each relation $\ttr$, we compute the average number of tails per head ($tphr$) and the average number of head per tail ($hptr$). If $tphr < 1.5$ and $hptr < 1.5$, $\ttr$ is treated as one-to-one; if $tphr \ge 1.5$ and $hptr \ge 1.5$, $\ttr$ is treated as a many-to-many; if $tphr < 1.5$ and $hptr \ge 1.5$, $\ttr$ is treated as one-to-many.}. 
The results of RotatE on different relation categories on the data set FB15k are summarized into Table~\ref{tab:one2one}. We also compare an additional approach KG2E\_KL \citep{he2015learning}, which is a probabilistic framework for knowledge graph embedding methods and aims to model the uncertainties of the entities and relations in knowledge graphs with the TransE model. We also summarize the statistics of different relation categories into Table~\ref{tab:one2onest} in the appendix.

We can see that besides the one-to-one relation, the RotatE model also performs quite well on the non-injective relations, especially on many-to-many relations. We also notice that the probabilistic framework KG2E\_KL(bern) \citep{he2015learning} is quite powerful, which consistently outperforms its corresponding knowledge graph embedding model, showing the importance of modeling the uncertainties in knowledge graphs. We leave the work of modeling the uncertainties in knowledge graphs with RotatE as our future work.


\section{Conclusion}
We have proposed a new knowledge graph embedding method called \method{}, which represents entities as complex vectors and relations as rotations in complex vector space. In addition, we propose a novel self-adversarial negative sampling technique for efficiently and effectively training the RotatE model. Our experimental results show that the \method{} model outperforms all existing state-of-the-art models on four large-scale benchmarks. Moreover, \method{} also achieves state-of-the-art results on a benchmark that is explicitly designed for composition pattern inference and modeling. A deep investigation into \method{} relation embeddings shows that the three relation patterns are implicitly represented in the relation embeddings. In the future, we plan to evaluate the \method{} model on more datasets and leverage a probabilistic framework to model the uncertainties of entities and relations.

\bibliography{iclr2019_conference}
\bibliographystyle{iclr2019_conference}

\newpage
\appendix

\section*{Appendix}

\section{Discussion on the Ability of Pattern Modeling and Inference} \label{sec:dis}
No existing models are capable of modeling all the three relation patterns. For example, TransE cannot model the symmetry pattern because it would yield $\rvr = \mathbf{0}$ for symmetric relations; TransX can infer and model the symmetry/antisymmetry pattern when $g_{r, 1} = g_{r, 2}$, e.g. in TransH  \citep{wang2014knowledge}, but cannot infer inversion and composition as $g_{r, 1}$ and $g_{r, 2}$ are invertible matrix multiplications; due to its symmetric nature, DistMult is difficult to model the asymmetric and inversion pattern; ComplEx addresses the problem of DisMult and is able to infer both the symmetry and asymmetric patterns with complex embeddings. Moreover, it can infer inversion rules because the complex conjugate of the solution to $\argmax_{\rvr}\Re(\langle \rvx, \rvr, \overline{\rvy} \rangle)$ is exactly the solution to $\argmax_{\rvr} \Re(\langle \rvy, \rvr, \overline{\rvx} \rangle)$. However, ComplEx cannot infer composition rules, since it does not model a bijection mapping from $\rvh$ to $\rvt$ via relation $\rvr$. These concerns are summarized in Table \ref{tab:modeling}.

\section{Proof of Lemma \ref{lem1}} \label{sec:lem1}
\begin{proof}
if $\ttr(\ttx, \tty)$ and $\ttr(\tty, \ttx)$ hold, we have
\begin{equation*}
	\rvy = \rvr \circ \rvx \wedge \rvx = \rvr \circ \rvy \Rightarrow \rvr\circ\rvr = \mathbf{1}
\end{equation*}
Otherwise, if $\ttr(\ttx, \tty)$ and $\neg \ttr(\tty, \ttx)$ hold, we have
\begin{equation*}
	\rvy = \rvr \circ \rvx \wedge \rvx \ne \rvr \circ \rvy \Rightarrow \rvr\circ\rvr \ne \mathbf{1} \qedhere
\end{equation*}
\end{proof}
\section{Proof of Lemma \ref{lem2}} \label{sec:lem2}
\begin{proof}
if $\ttr_1(\ttx, \tty)$ and $\ttr_2(\tty, \ttx)$ hold, we have
\begin{equation*}
	\rvy = \rvr_1 \circ \rvx \wedge \rvx = \rvr_2 \circ \rvy \Rightarrow \rvr_1 = \rvr_2^{-1} \qedhere
\end{equation*}
\end{proof}
\section{Proof of Lemma \ref{lem3}} \label{sec:lem3}
\begin{proof}
if $\ttr_1(\ttx, \ttz)$, $\ttr_2(\ttx, \tty)$ and $\ttr_3(\tty, \ttz)$ hold, we have
\begin{equation*}
	\rvz = \rvr_1 \circ \rvx \wedge \rvy = \rvr_2 \circ \rvx \wedge \rvz = \rvr_3 \circ \rvy \Rightarrow \rvr_1 = \rvr_2 \circ \rvr_3 \qedhere
\end{equation*}
\end{proof}

\section{Properties of \method{}} A useful property for \method{} is that the inverse of a relation can be easily acquired by complex conjugate. In this way, the \method{} model treats head and tail entities in a uniform way, which is potentially useful for efficient 1-N scoring \citep{dettmers2017convolutional}:
\begin{equation}
	\norm{\rvh \circ \rvr - \rvt} = \norm{(\rvh \circ \rvr - \rvt) \circ \overline{\rvr}} = \norm{\rvt \circ \overline{\rvr} - \rvh}
\end{equation}

Moreover, considering the embeddings in the polar form, i.e., $h_i = m_{h, i} e^{i \theta_{h, i}}, r_i = e^{i \theta_{r, i}}, t_i = m_{t, i} e^{i \theta_{t, i}}$, we can rewrite the \method{} distance function as:
\begin{equation}\label{eqa:pro}
\norm{\rvh \circ \rvr - \rvt} = \sum_{i=1}^k \sqrt{ (m_{h, i} - m_{t, i})^2 + 4 m_{h, i}m_{t, i} \sin^2 \frac{\theta_{h, i} +  \theta_{r, i} -  \theta_{t, i}}{2}}
\end{equation}

This equation provides two interesting views of the model:

(1) When we constrain the modulus $m_{h, i} = m_{t, i} = C$, the distance function is reduced to $2C\norm{\sin\frac{\boldsymbol{\theta}_h + \boldsymbol{\theta}_r - \boldsymbol{\theta}_t}{2}}$. We can see that this is very similar to the distance function of TransE: $\norm{\rvh+\rvr-\rvt}$. Based on this intuition, we can show that:
\begin{theorem}\label{the1}
\method{} can degenerate into TransE. (See proof at Appendix \ref{sec:the1})
\end{theorem}
which indicates that \method{} is able to simulate TransE.

(2) The modulus provides the lower bound of the distance function, which is $\norm{\rvm_h - \rvm_t}$.

\section{Proof of Theorem \ref{the1}}  \label{sec:the1}
\begin{proof}
By further restricting $|h_i| = |t_i| = C$, we can rewrite $\rvh, \rvr, \rvt$ by
\begin{align}
\rvh &= Ce^{i\boldsymbol{\theta}_h} = C\cos \boldsymbol{\theta}_h + i C\sin\boldsymbol{\theta}_h\\
\rvr &= e^{i\boldsymbol{\theta}_r} = \cos \boldsymbol{\theta}_r + i \sin\boldsymbol{\theta}_r\\
\rvt &= Ce^{i\boldsymbol{\theta}_t} = C\cos \boldsymbol{\theta}_t + i C\sin\boldsymbol{\theta}_t\\
\end{align}
Therefore, we have
\begin{align}
\norm{\rvh \circ \rvr - \rvt} 
&= C\norm{e^{i(\boldsymbol{\theta}_h +\boldsymbol{\theta}_r)} - e^{i\boldsymbol{\theta}_t}} = C \norm{e^{i(\boldsymbol{\theta}_h +\boldsymbol{\theta}_r - \boldsymbol{\theta}_t)} - \mathbf{1}}\\
&= C\norm{\cos (\boldsymbol{\theta}_h +\boldsymbol{\theta}_r - \boldsymbol{\theta}_t) - \mathbf{1} + i \sin (\boldsymbol{\theta}_h +\boldsymbol{\theta}_r - \boldsymbol{\theta}_t)}\\
&= C\norm{\sqrt{(\cos (\boldsymbol{\theta}_h +\boldsymbol{\theta}_r - \boldsymbol{\theta}_t) - \mathbf{1})^2 + \sin^2 (\boldsymbol{\theta}_h +\boldsymbol{\theta}_r - \boldsymbol{\theta}_t)}}\\
&= C\norm{\sqrt{\mathbf{2} - 2\cos (\boldsymbol{\theta}_h +\boldsymbol{\theta}_r - \boldsymbol{\theta}_t)}}\\
&= 2C\norm{\sin \frac{\boldsymbol{\theta}_h +\boldsymbol{\theta}_r - \boldsymbol{\theta}_t}{2}} \label{eqa:baseline}
\end{align}
If the embedding of $(\tth, \ttr, \ttt)$ in TransE is $\rvh', \rvr', \rvt'$, let $\boldsymbol{\theta}_h = c\rvh', \boldsymbol{\theta}_r = c\rvr', \boldsymbol{\theta}_t = c\rvt'$ and $C = 1/c$ , we have
\begin{equation*}
\lim_{c \rightarrow 0}\norm{\rvh \circ \rvr - \rvt} = \norm{\rvh' + \rvr' - \rvt'} \qedhere
\end{equation*}
\end{proof}

\begin{table}[t]
\centering
\small
\begin{tabular}{|c|c|c|c|c|}
\hline
Relation Category & 1-to-1 & 1-to-N & N-to-1 & N-to-N\\
\hline
\#relation & 326 & 308 & 388 & 323\\
\hline
\#triplet (train) & 6827 & 42509 & 70727 & 363079\\
\hline
\#triplet (test) & 832 & 5259 & 8637 & 44343\\
\hline
\end{tabular}
\caption{Statistics of FB15k by mapping properties of relations.}
\label{tab:one2onest}
\end{table}

\begin{table}[t]
\centering
\begin{tabular}{|c|c c c c c|}
\hline
& \multicolumn{5}{c|}{\textbf{YAGO3-10}}\\
\cline{2-6}
& MR & MRR & H@1 & H@3 & H@10\\
\hline
DistMult & 5926 & .34 & .24 & .38 & .54\\
\hline
ComplEx & 6351 & .36 & .26 & .40 & .55\\
\hline
ConvE & \textbf{1671} & .44 & .35 & .49 & .62\\
\hline
\hline
\method{} & 1767 & \textbf{.495} & \textbf{.402} & \textbf{.550} & \textbf{.670} \\
\hline
\end{tabular}
\caption{Results of several models evaluated on the YAGO3-10 datasets. Other results are taken from \citep{dettmers2017convolutional}.}
\label{tab:yago}
\end{table}

\section{Link Prediction on YAGO3-10}
YAGO3-10 is a subset of YAGO3 \citep{mahdisoltani2013yago3}, which consists of entities that have a minimum of 10 relations each. It has 123,182 entities and 37 relations. Most of the triples deal with descriptive attributes of people, such as citizenship, gender, profession and marital status.

Table \ref{tab:yago} shows that the \method{} model also outperforms state-of-the-art models on YAGO3-10.

\begin{table}[t]
\centering
\begin{tabular}{|c|c c c c c|}
\hline
Benchmark & embedding dimension $k$ & batch size $b$ & negative samples $n$ & $\alpha$ & $\gamma$ \\
\hline
FB15k & 1000 & 2048 & 128 & 1.0 & 24 \\
\hline
WN18 & 500 & 512 & 1024 & 0.5 & 12 \\
\hline
FB15k-237 & 1000 & 1024 & 256 & 1.0 & 9 \\
\hline
WN18RR & 500 & 512 & 1024 & 0.5 & 6 \\
\hline
Countries S1 & 500 & 512 & 64 & 1.0 & 0.1 \\
\hline
Countries S2 & 500 & 512 & 64 & 1.0 & 0.1 \\
\hline
Countries S3 & 500 & 512 & 64 & 1.0 & 0.1 \\
\hline
YAGO3-10 & 500 & 1024 & 400 & 1.0 & 24 \\
\hline
\end{tabular}
\caption{The best hyperparameter setting of \method{} on several benchmarks.}
\label{tab:hyper}
\end{table}

\section{Hyperparameters}
We list the best hyperparameter setting of \method{} w.r.t the validation dataset on several benchmarks in Table \ref{tab:hyper}.

\section{Ablation Study}
Table \ref{tab:ablation} shows our ablation study of self-adversarial sampling and negative sampling loss on FB15k-237. We also re-implement a 1000-dimension TransE and do ablation study on it. From the table, We can find that self-adversarial sampling boosts the performance for both models, while negative sampling loss is only effective on \method{}; in addition, our re-implementation of TransE also outperforms all the state-of-the-art models on FB15k-237.

\begin{table}[t]
\centering
\small
\begin{tabular}{|c|c c c c c|c c c c c|}
\hline
& \multicolumn{5}{c|}{\textbf{\method{}}} & \multicolumn{5}{c|}{\textbf{TransE}}\\
\cline{2-11}
& MR & MRR & H@1 & H@3 & H@10 & MR & MRR & H@1 & H@3 & H@10 \\
\hline
\multicolumn{11}{|c|}{negative sampling loss}\\
\hline
w/ adv & \textbf{177} & \textbf{.338} & \textbf{.241} & \textbf{.375} & \textbf{.533} & 170 & .332 & .233 & \textbf{.372} & \textbf{.531}\\
\hline
w/o adv & 185 & .297 & .205 & .328 & .480 & 175 & .297 & .202 & .331 & .486\\
\hline
\multicolumn{11}{|c|}{margin-based ranking criterion}\\
\hline
w/ adv &  225 & .322 & .225 & .358 & .516 & 167 & \textbf{.333} & \textbf{.237} & .370 & .522\\
\hline
w/o adv & 199 & .293 & .202 & .324 & .476 & \textbf{164} & .306 & .212 & .340 & .493\\
\hline
\end{tabular}
\caption{Results of ablation study on FB15k-237, where ``adv'' represents ``self-adversarial''.}
\label{tab:ablation}
\end{table}

\begin{table}[t]
\centering
\small
\begin{tabular}{|c|c|c|c|c|}
\hline
& FB15k & WN18 & FB15k-237 & WN18RR \\
\hline
MRR & $.797 \pm .001$ & $.949 \pm .000$ & $.337 \pm .001$ & $.477 \pm .001$\\
\hline
\end{tabular}
\caption{The average and variance of the MRR results of RotatE on FB15k, WN18, FB15k-237 and WN18RR.}
\label{tab:var}
\end{table}

\section{Variance of the Results}

In Table \ref{tab:var}, We provide the average and variance of the MRR results on FB15k, WN18, FB15k-237 and WN18RR. Both the average and the variance is calculated by three runs of RotatE with difference random seeds. We can find that the performance of RotatE is quite stable for different random initialization.

\section{More results of implicit basic pattern inference}

\begin{figure}[t]
\centering
\begin{subfigure}{0.45\textwidth}
  \centering
  \includegraphics[width=\linewidth]{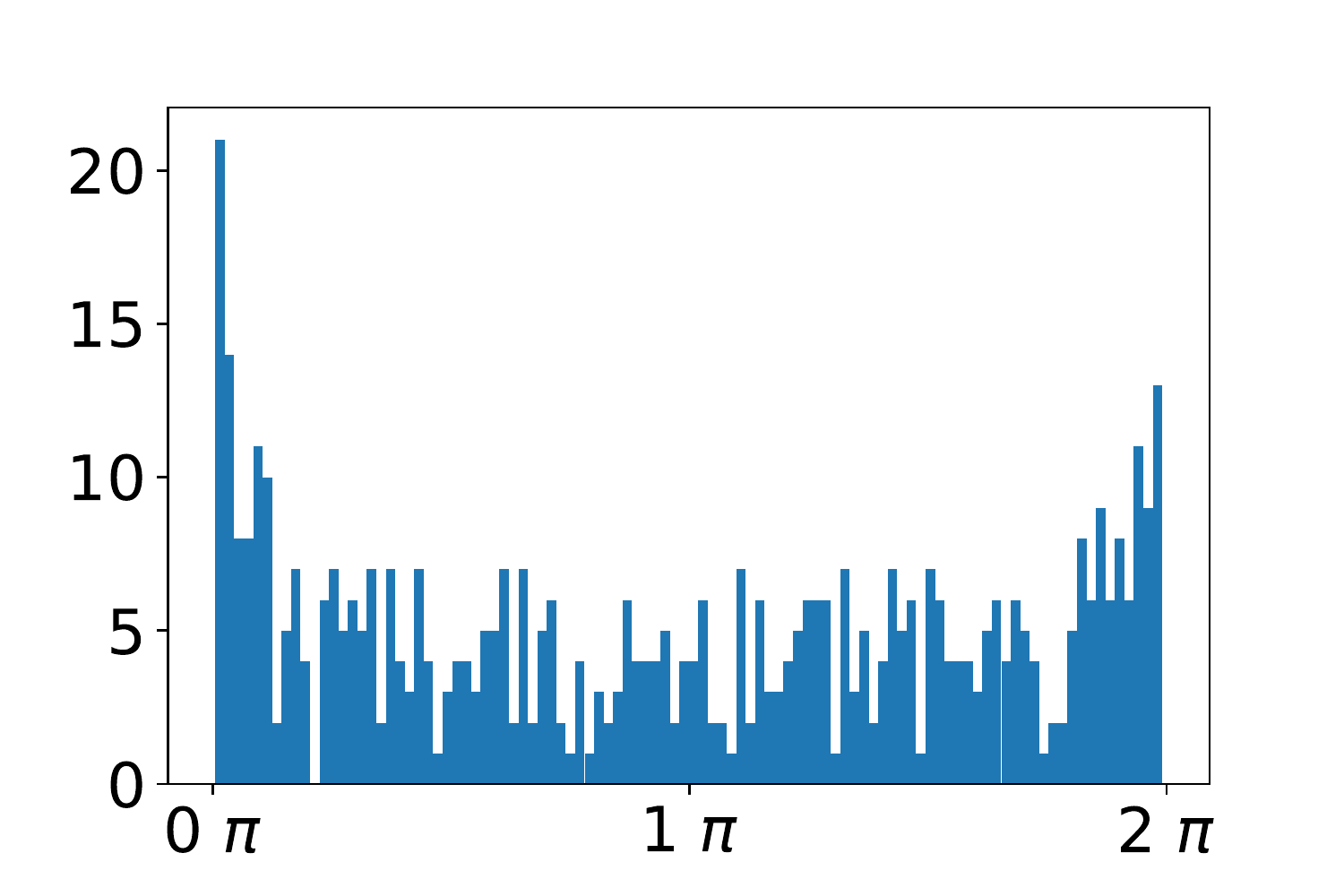}
  \caption{$\mathtt{has\_part}$}
\end{subfigure}%
\begin{subfigure}{0.45\textwidth}
  \centering
  \includegraphics[width=\linewidth]{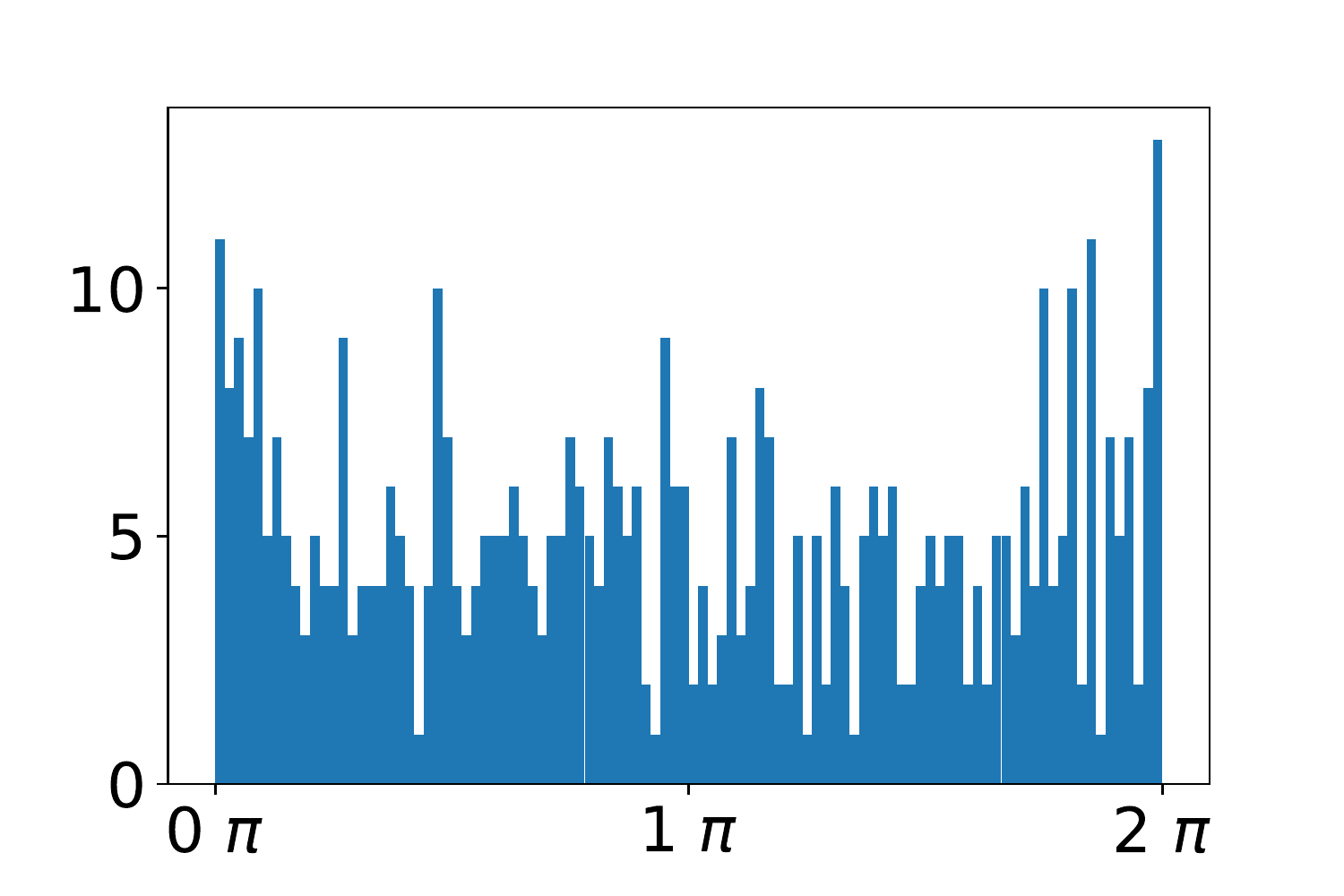}
  \caption{$\mathtt{instance\_hypernym}$}
\end{subfigure}
\begin{subfigure}{0.45\textwidth}
  \centering
  \includegraphics[width=\linewidth]{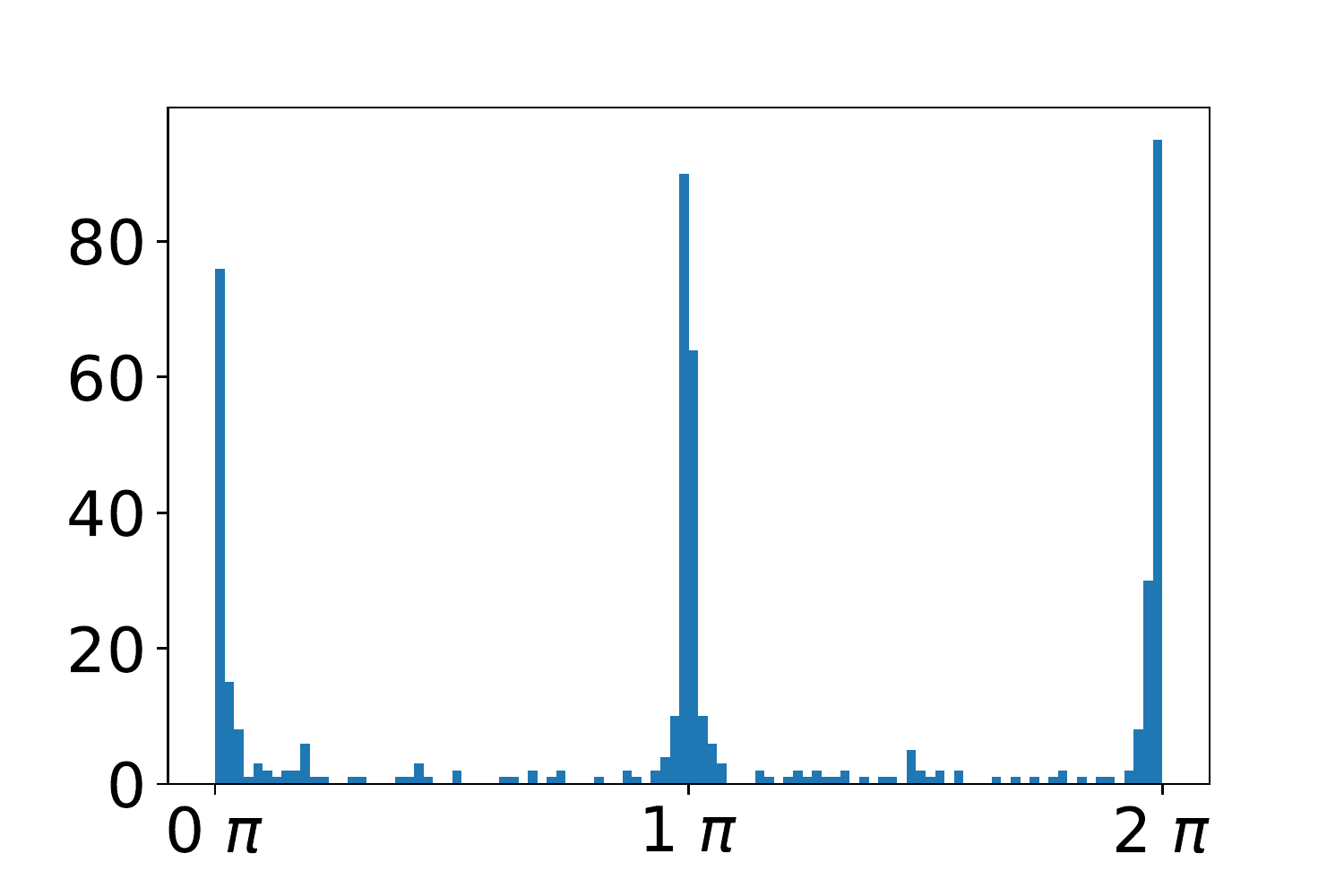}
  \caption{$\mathtt{also\_see}$}
\end{subfigure}
\begin{subfigure}{0.45\textwidth}
  \centering
  \includegraphics[width=\linewidth]{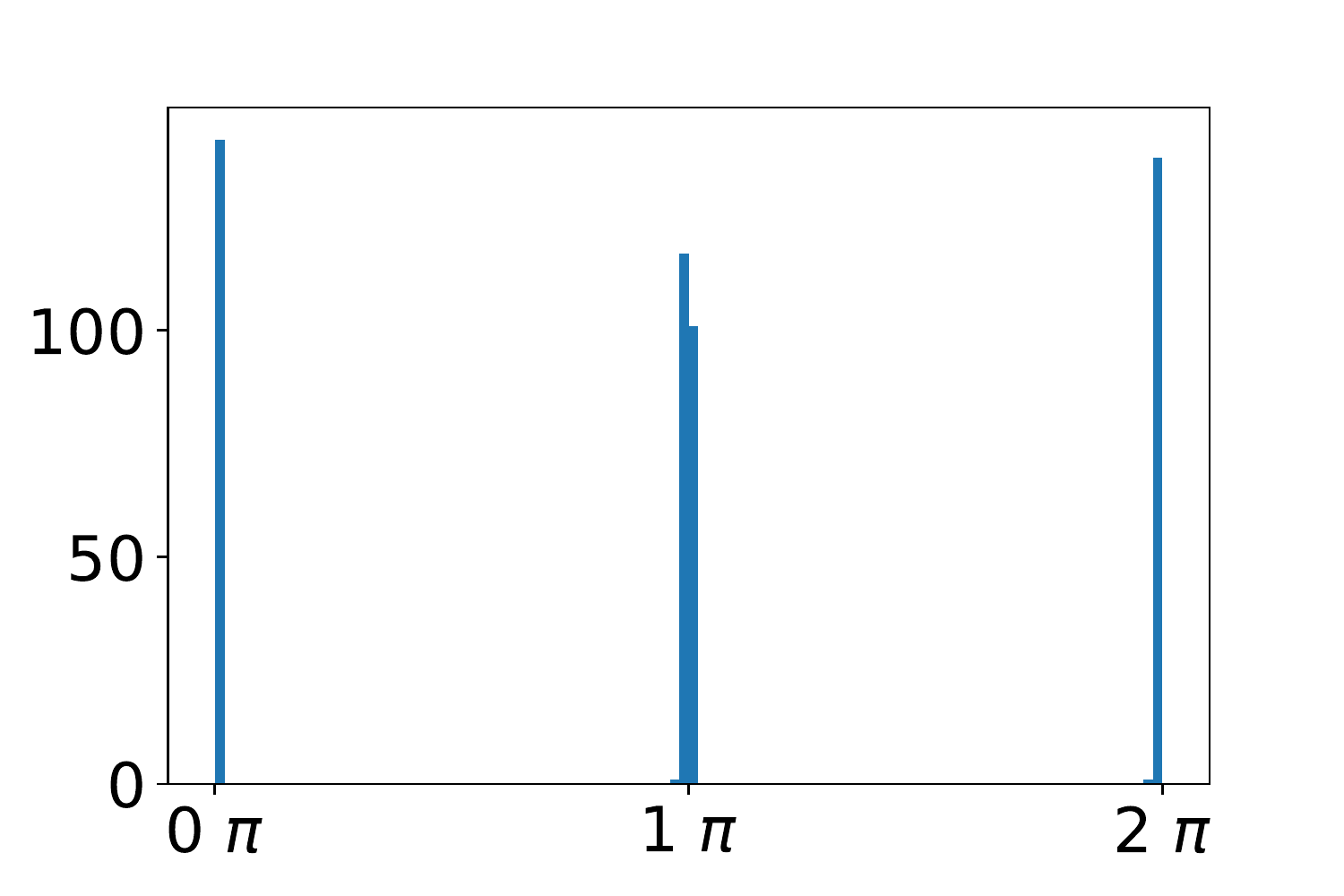}
  \caption{$\mathtt{verb\_group}$}
\end{subfigure}
\begin{subfigure}{0.45\textwidth}
  \centering
  \includegraphics[width=\linewidth]{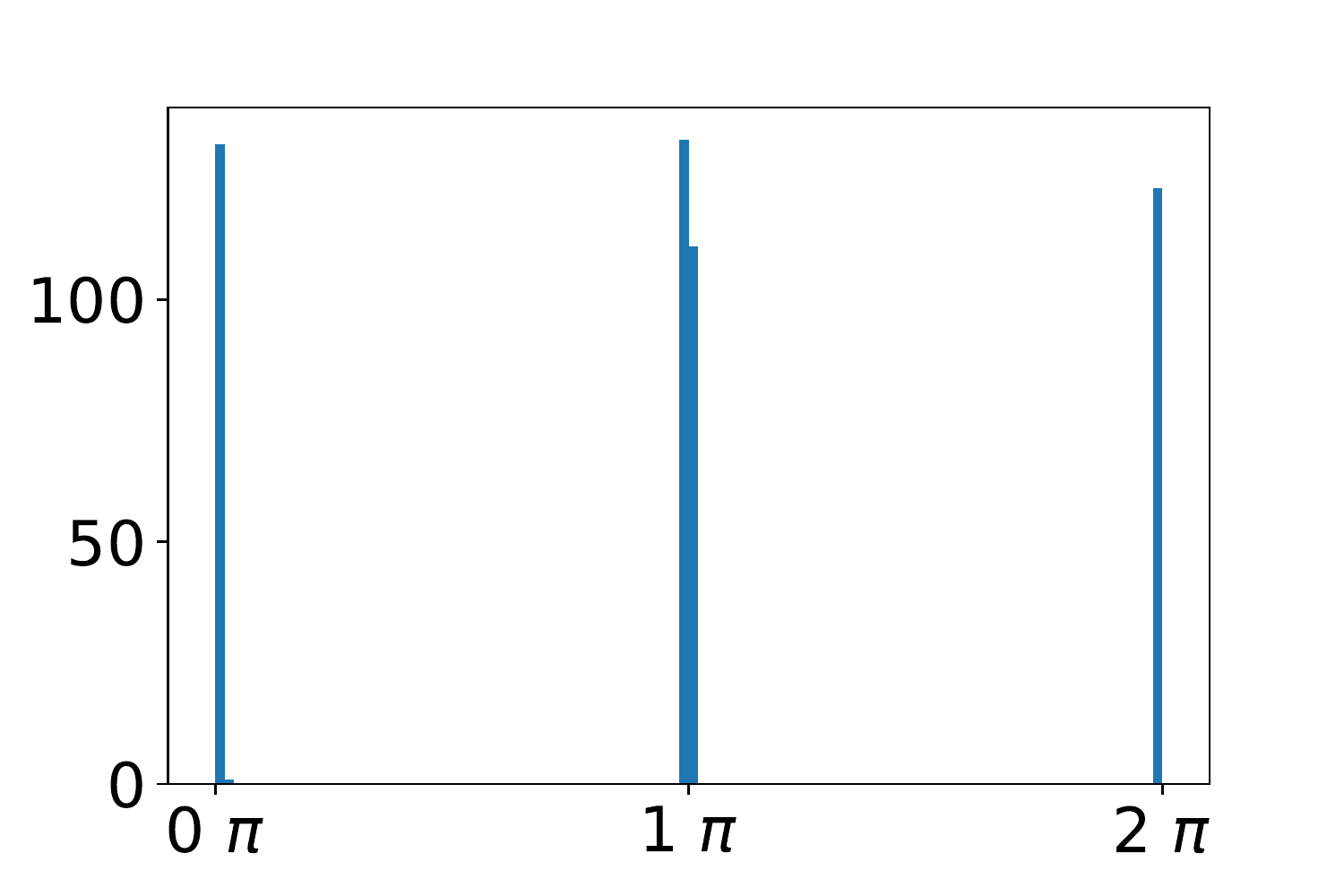}
  \caption{$\mathtt{derivationally\_related\_form}$}
\end{subfigure}
\begin{subfigure}{0.45\textwidth}
  \centering
  \includegraphics[width=\linewidth]{11-eps-converted-to.pdf}
  \caption{$\mathtt{similar\_to}$}
\end{subfigure}
\caption{Histograms of embedding phases from two general relations and four symmetric relations on WN18. ( $k = 500$ )}
\label{fig:wn1}
\end{figure}

\begin{figure}
\centering
\begin{subfigure}{0.45\textwidth}
  \centering
  \includegraphics[width=\linewidth]{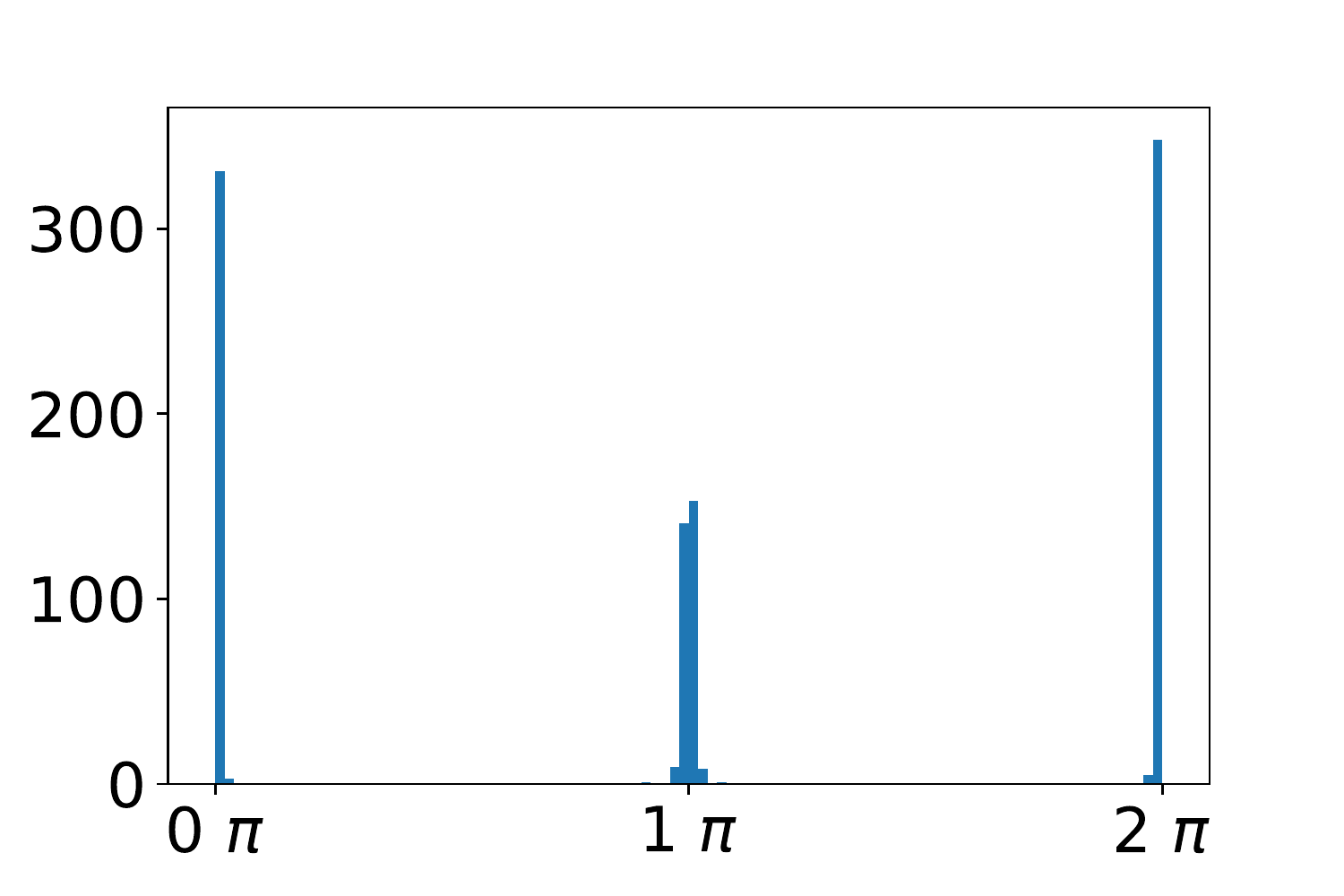}
  \caption{$\mathtt{/celebrities /celebrity} $-\\ $\mathtt{/celebrity\_friends./celebrities}$- $\mathtt{/friendship /friend}$}
\end{subfigure}%
\begin{subfigure}{0.45\textwidth}
  \centering
  \includegraphics[width=\linewidth]{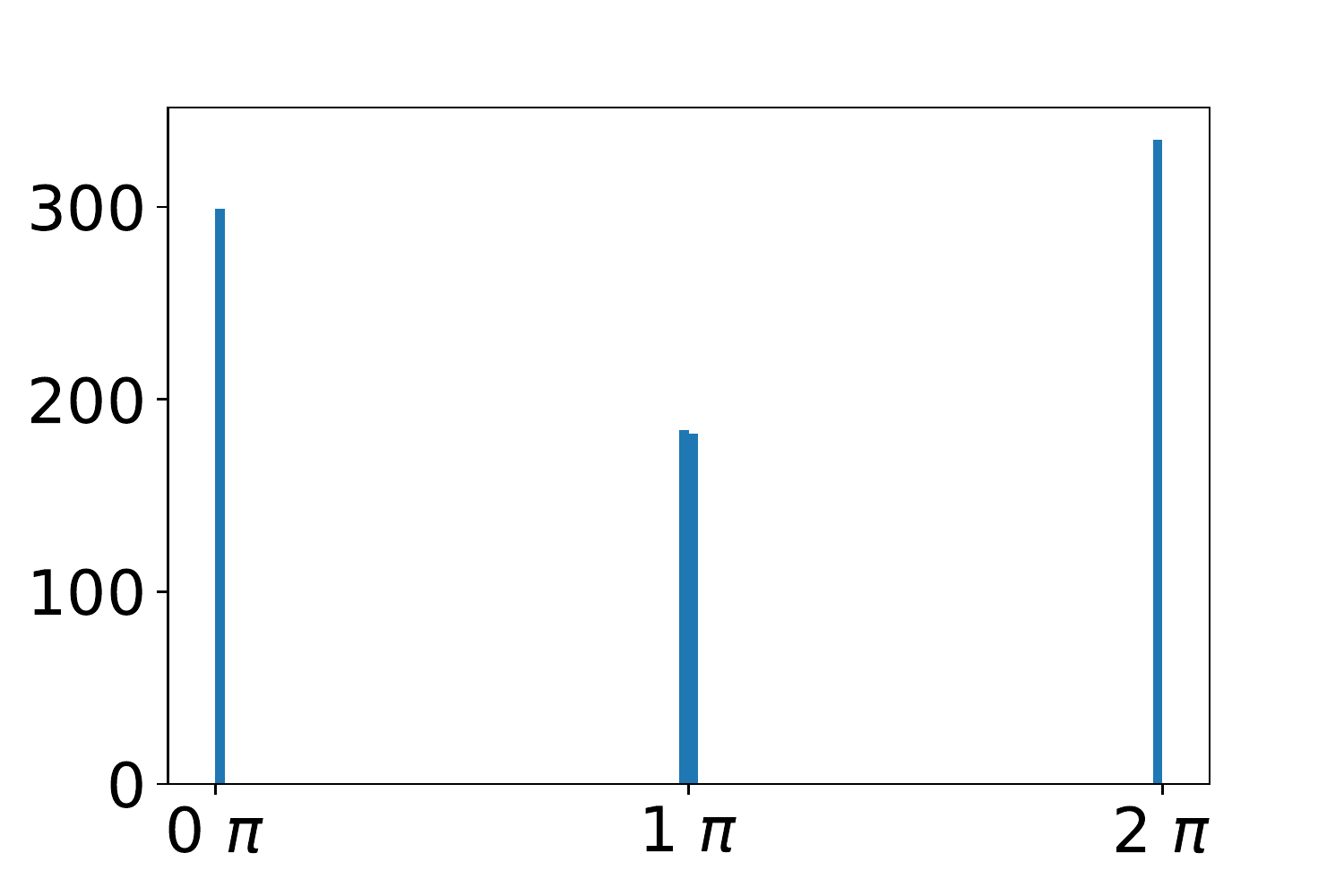}
  \caption{$\mathtt{/award/award\_winner }$-\\ $\mathtt{ /awards\_won./award/award\_honor}$- $\mathtt{/award\_winner}$}
\end{subfigure}
\begin{subfigure}{0.45\textwidth}
  \centering
  \includegraphics[width=\linewidth]{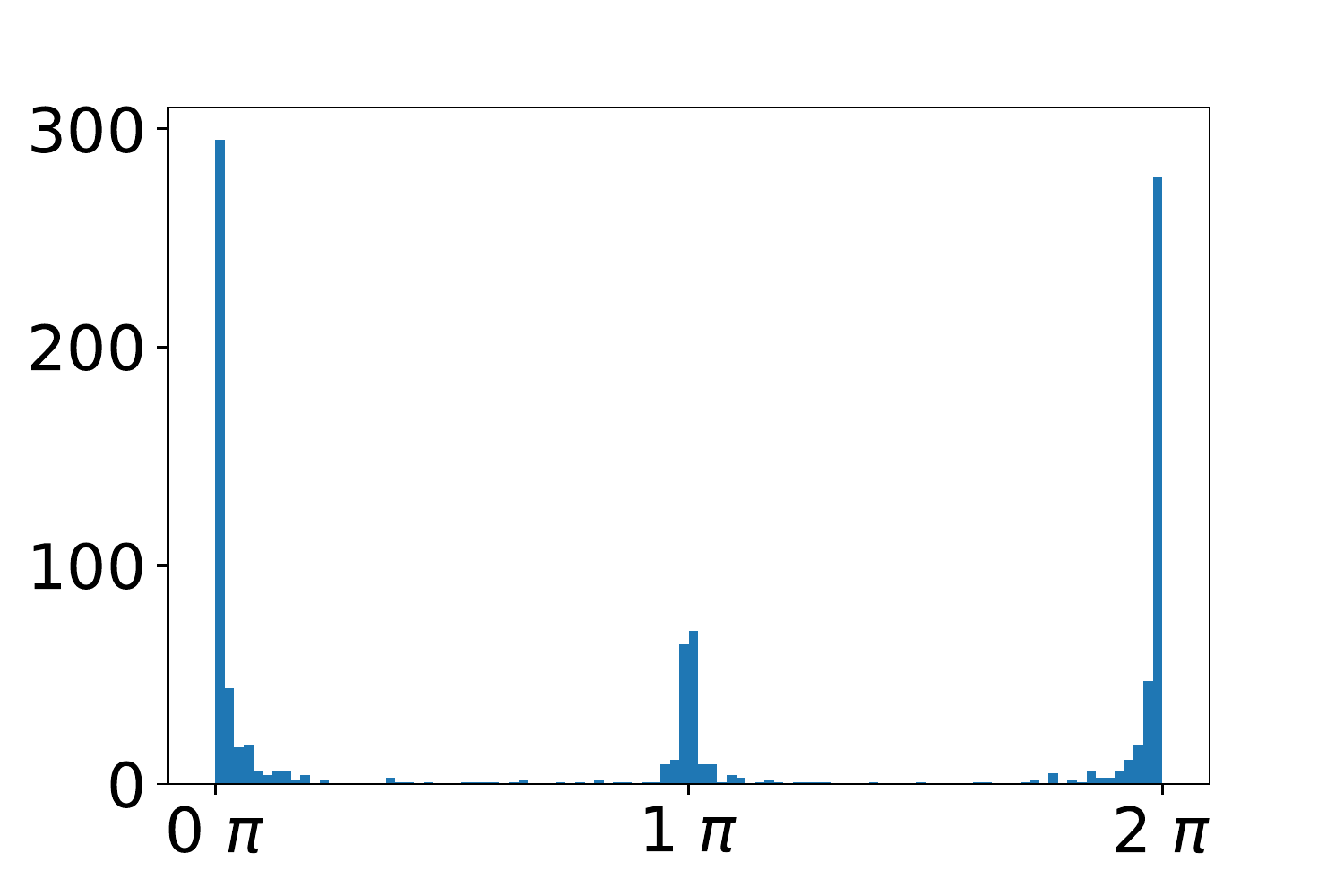}
  \caption{$\mathtt{/location/statistical\_region}$-\\$\mathtt{/places\_exported\_to./location}$- $\mathtt{/imports\_and\_exports/exported\_to}$}
\end{subfigure}
\begin{subfigure}{0.45\textwidth}
  \centering
  \includegraphics[width=\linewidth]{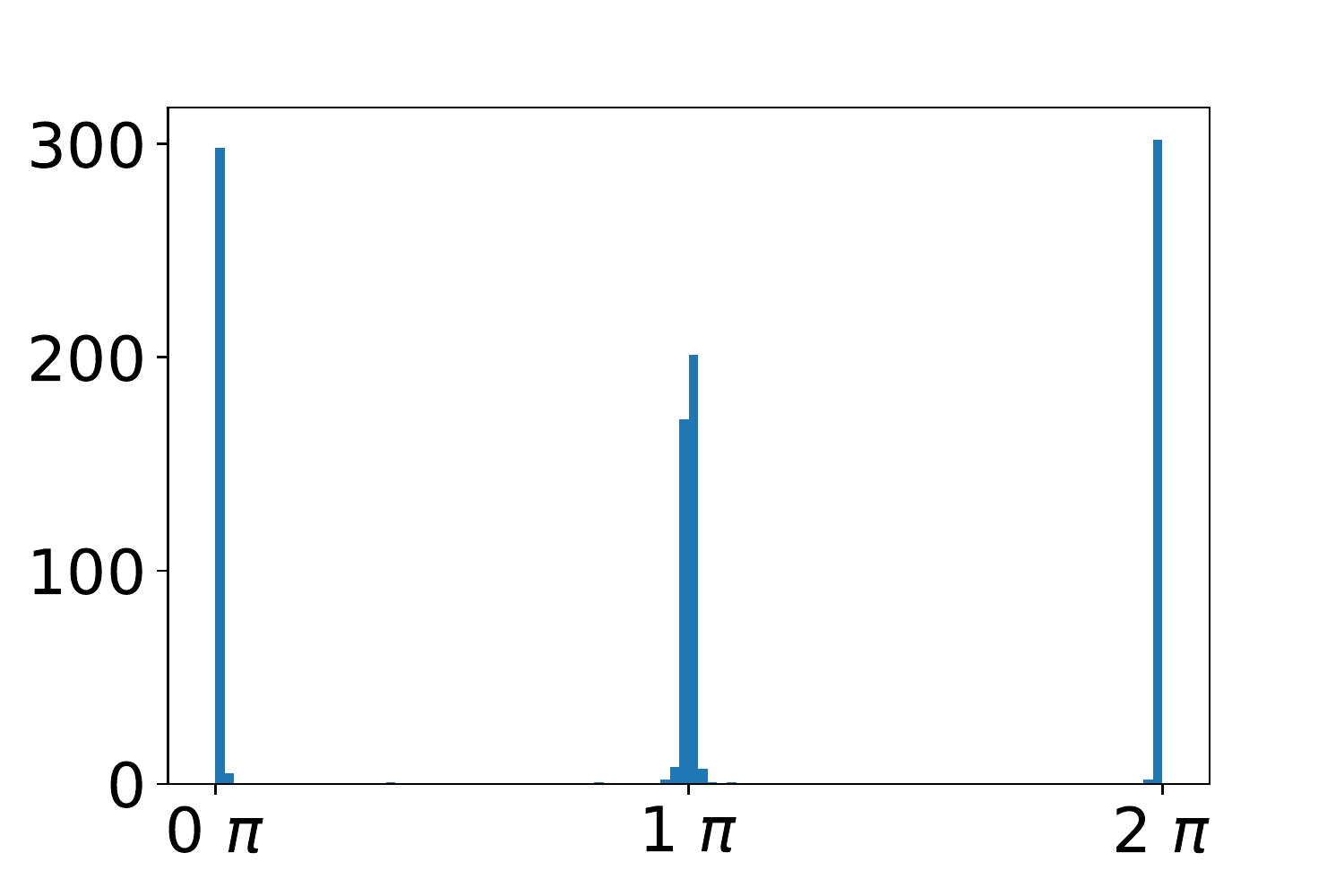}
  \caption{$\mathtt{/base/popstra/celebrity}$-\\ $\mathtt{/breakup./base/popstra}$- $\mathtt{/breakup/participant}$}
\end{subfigure}
\begin{subfigure}{0.45\textwidth}
  \centering
  \includegraphics[width=\linewidth]{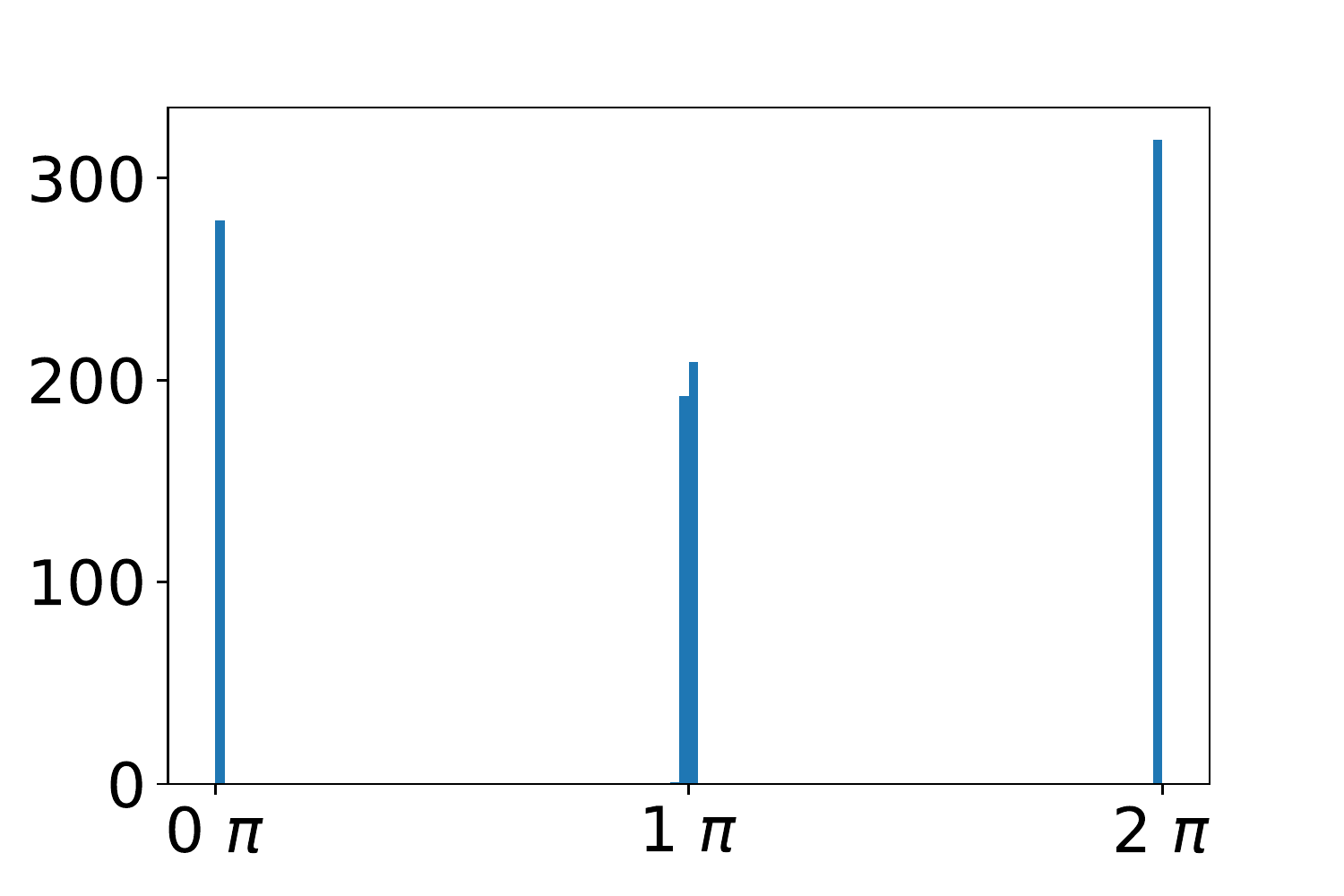}
  \caption{$\mathtt{/base/popstra/celebrity}$-\\ $\mathtt{/dated./base/popstra}$-\\ $\mathtt{/dated/participant}$}
\end{subfigure}
\begin{subfigure}{0.45\textwidth}
  \centering
  \includegraphics[width=\linewidth]{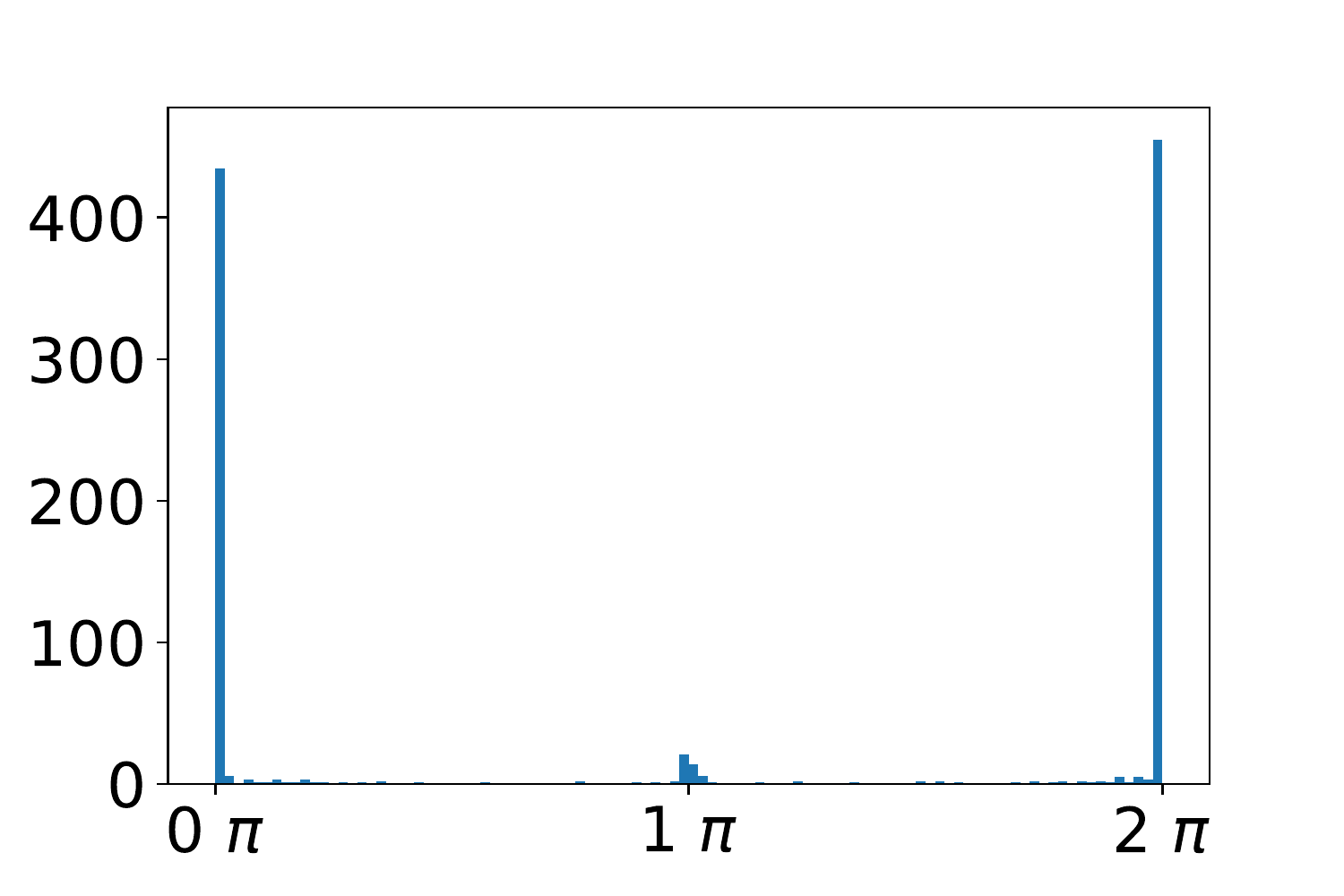}
  \caption{$\mathtt{/government/legislative\_session}$-\\ $\mathtt{/members./government}$- $\mathtt{/government\_position\_held/legislative\_sessions}$}
\end{subfigure}
\caption{Histograms of embedding phases from six symmetric relations on FB15k-237. ($k = 1000$)}
\label{fig:fb1}
\end{figure}

\begin{figure}
\centering
\begin{subfigure}{0.45\textwidth}
  \centering
  \includegraphics[width=\linewidth]{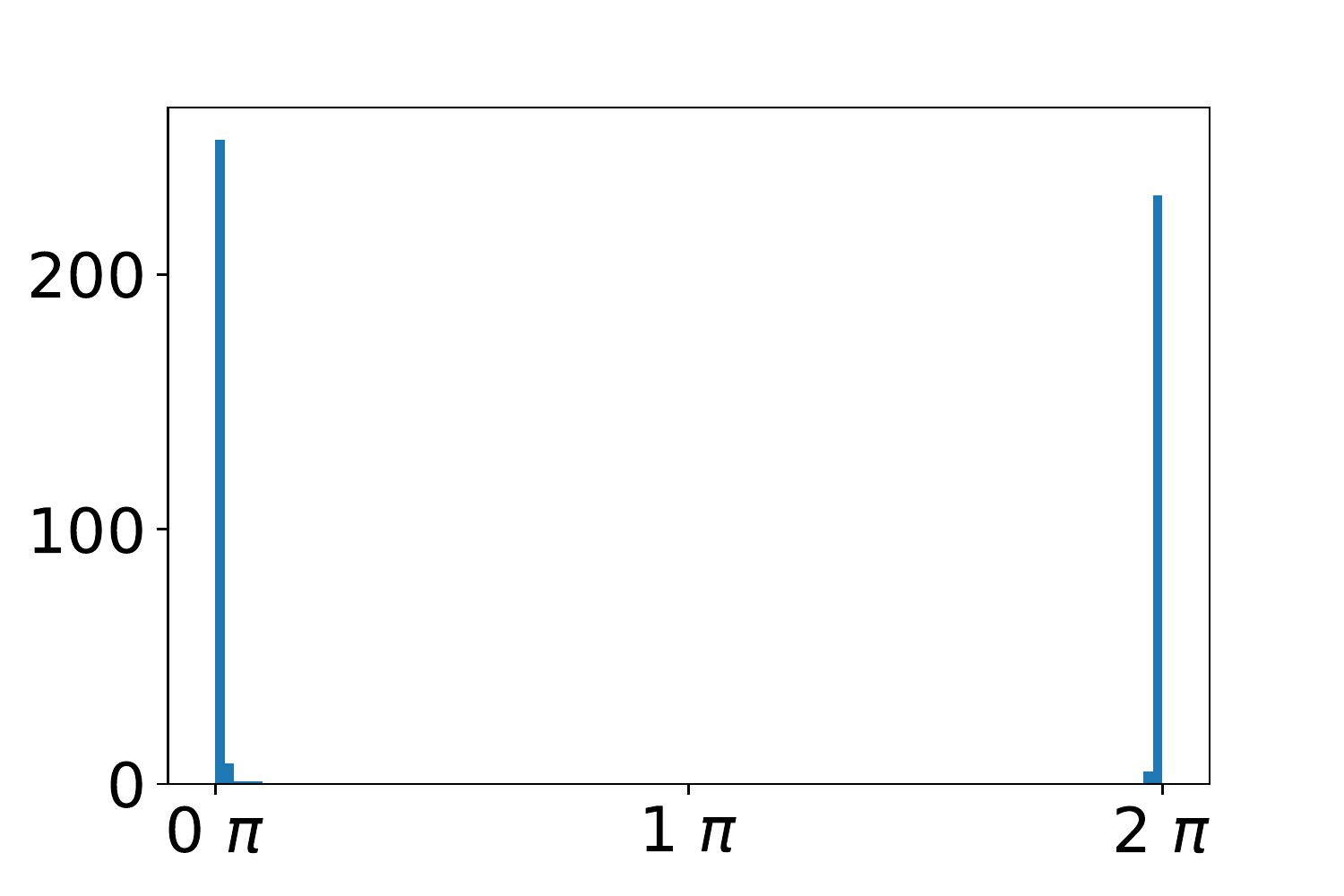}
  \caption{$\mathtt{member\_of\_domain\_topic}$\\$\mathtt{\circ\  synset\_domain\_topic\_of}$}
\end{subfigure}%
\begin{subfigure}{0.45\textwidth}
  \centering
  \includegraphics[width=\linewidth]{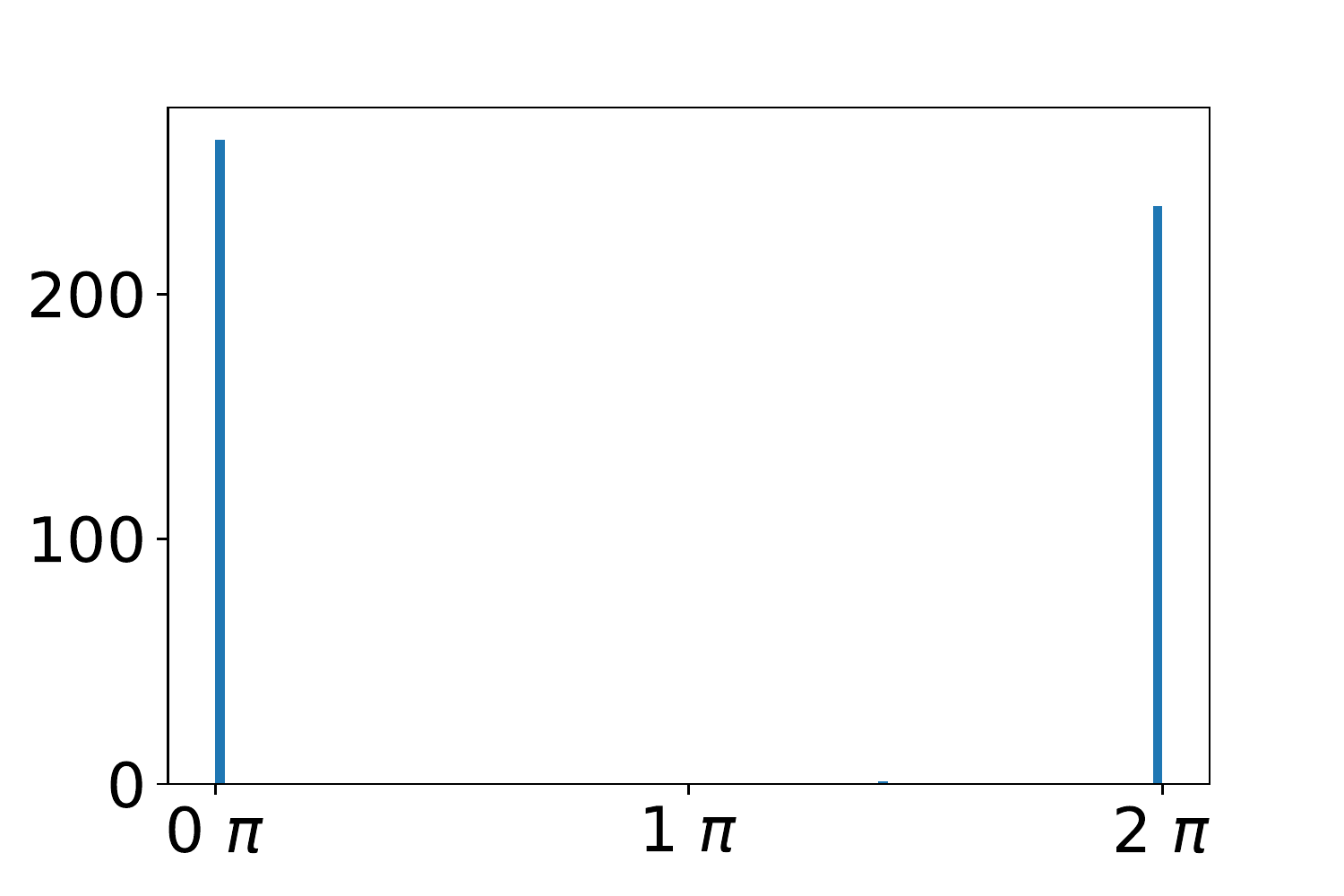}
  \caption{$\mathtt{has\_part \circ  part\_of}$}
\end{subfigure}
\begin{subfigure}{0.45\textwidth}
  \centering
  \includegraphics[width=\linewidth]{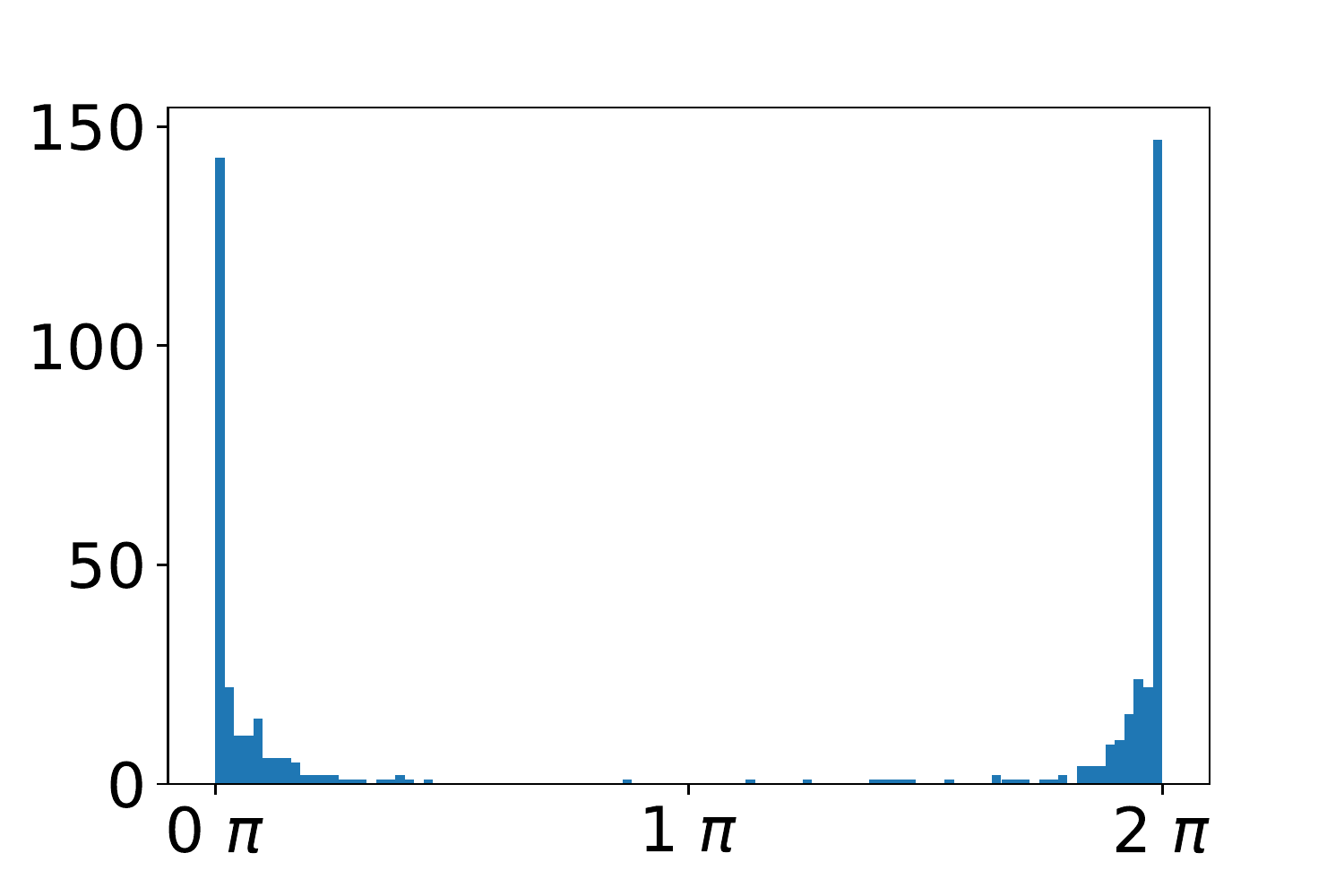}
  \caption{$\mathtt{synset\_domain\_usage\_of}$\\$\mathtt{\circ\  member\_of\_domain\_usage}$}
\end{subfigure}
\begin{subfigure}{0.45\textwidth}
  \centering
  \includegraphics[width=\linewidth]{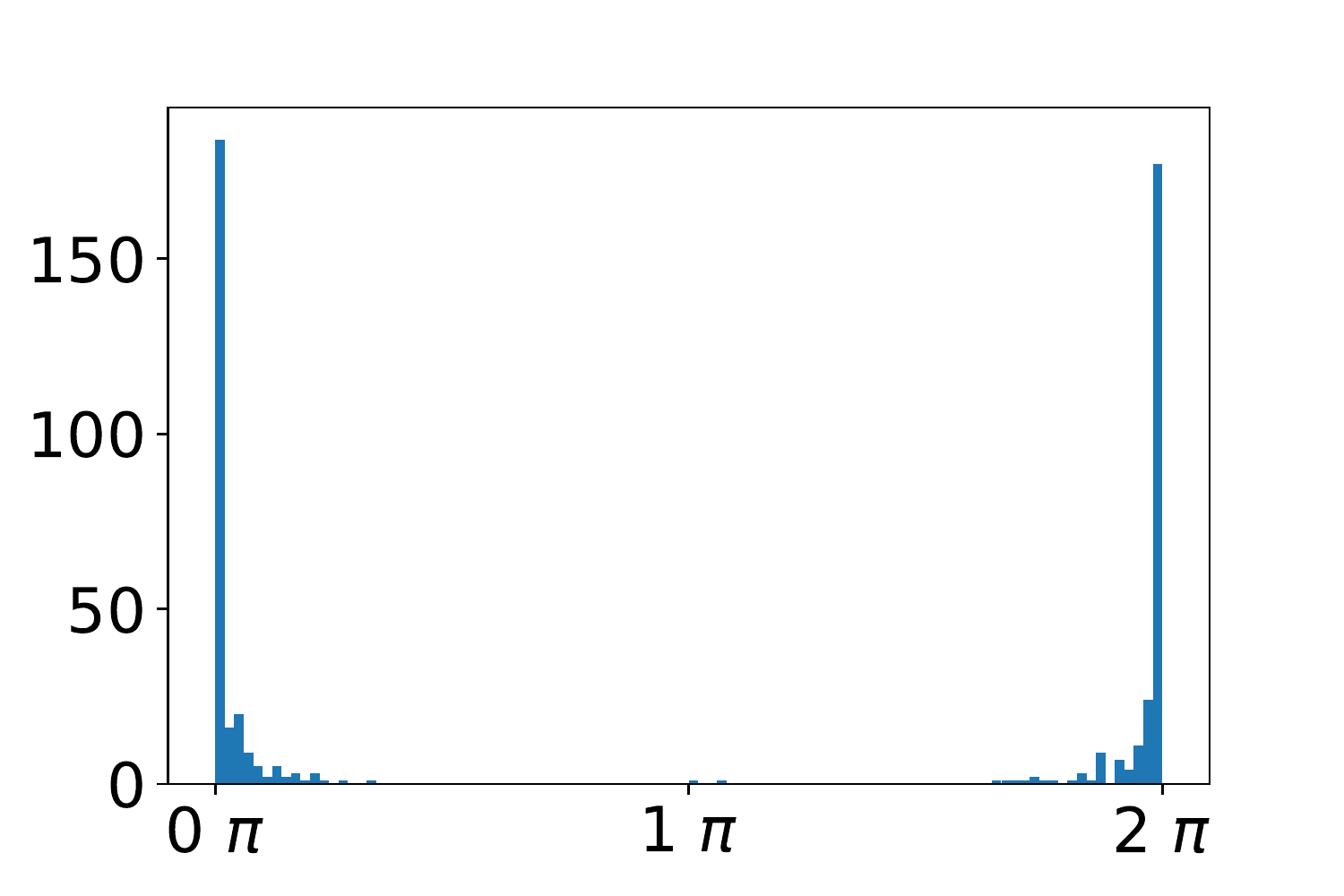}
  \caption{$\mathtt{synset\_domain\_region\_of}$\\$\mathtt{\circ\  member\_of\_domain\_region}$}
\end{subfigure}
\begin{subfigure}{0.45\textwidth}
  \centering
  \includegraphics[width=\linewidth]{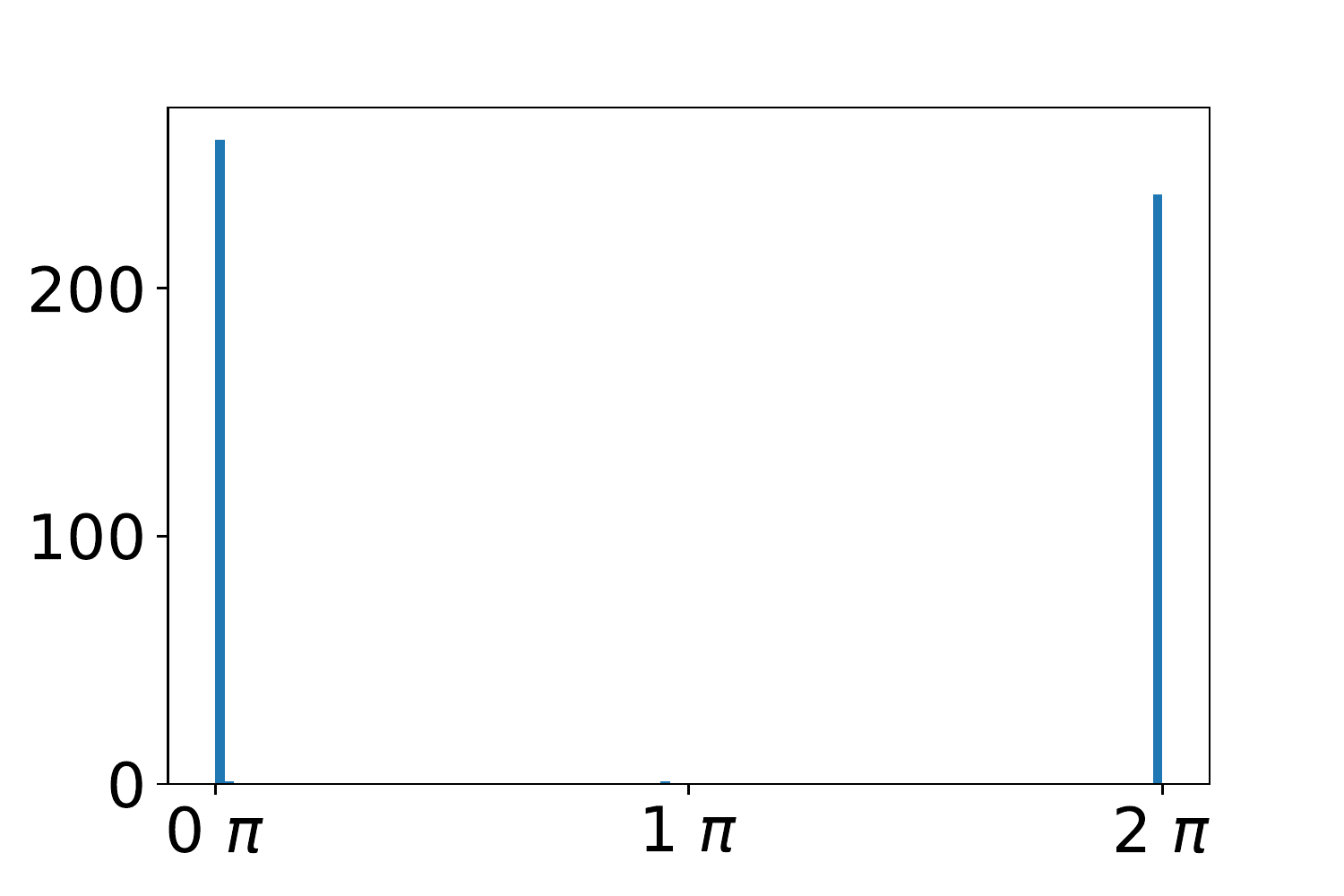}
  \caption{$\mathtt{member\_meronym \circ member\_holonym}$}
\end{subfigure}
\begin{subfigure}{0.45\textwidth}
  \centering
  \includegraphics[width=\linewidth]{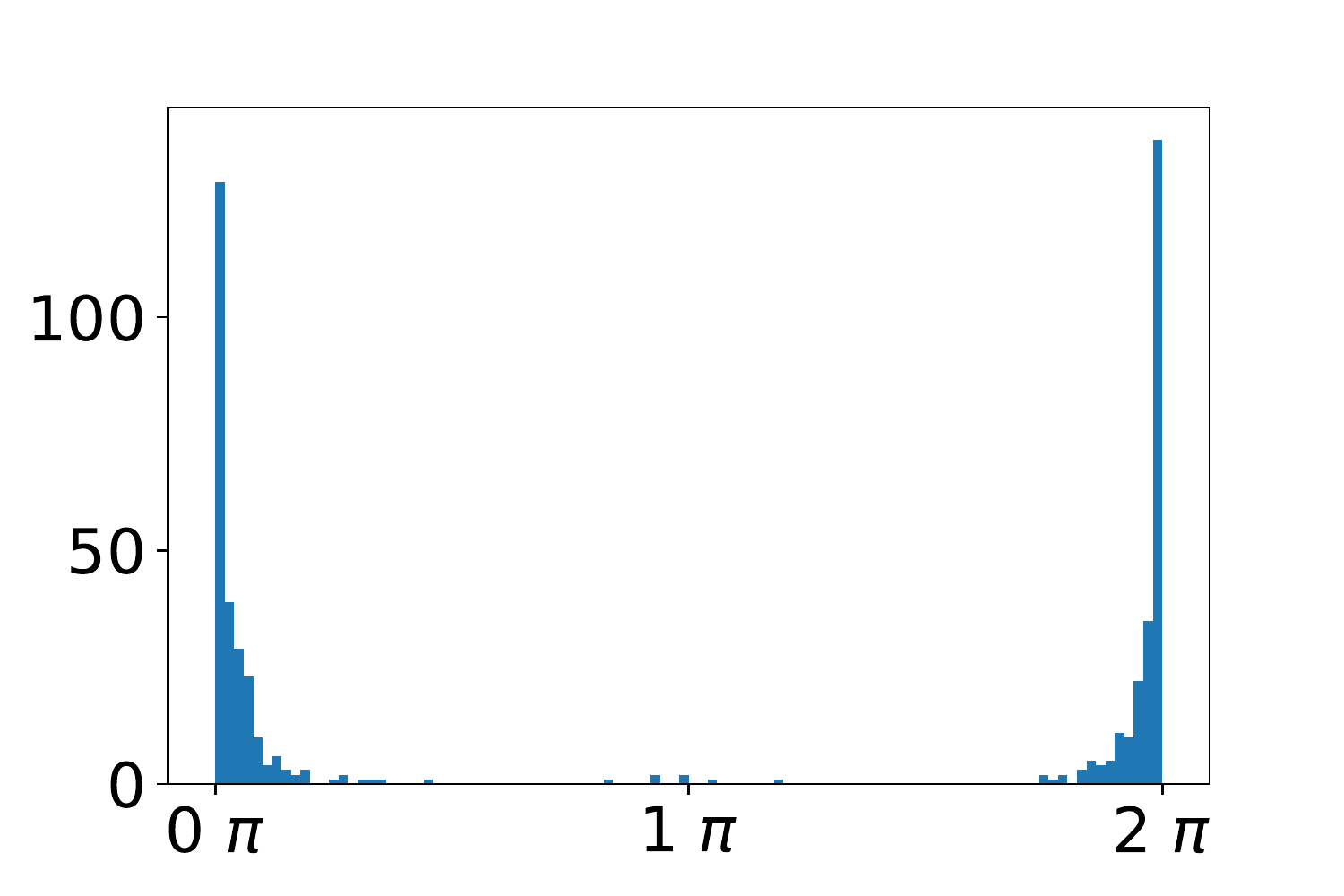}
  \caption{$\mathtt{instance\_hypernym \circ instance\_hyponym}$}
\end{subfigure}
\caption{Histograms of element-wise additions of inversed relation embedding phases on WN18. ($k = 500$)}
\label{fig:wn2}
\end{figure}

We provide more histograms of embedding phases in Figure \ref{fig:wn1} - \ref{fig:wn2}.

\newpage

\end{document}